\documentclass[letterpaper]{article} 
\usepackage{aaai2027}  
\usepackage[hyphens]{url}  
\usepackage{graphicx} 
\usepackage{natbib}  
\usepackage{caption} 
\nocopyright
\usepackage{amssymb}
\usepackage{multirow}
\usepackage{algorithm}
\usepackage{algorithmic}
\usepackage{amsmath}
\usepackage{makecell}
\usepackage[table]{xcolor}
\usepackage{rotating}

\usepackage{newfloat}
\usepackage{listings}
\DeclareCaptionStyle{ruled}{labelfont=normalfont,labelsep=colon,strut=off} 
\floatstyle{ruled}
\newfloat{listing}{tb}{lst}{}
\floatname{listing}{Listing}

\usepackage{booktabs}

\title{Thinking on Shots: Consistent Multi-Shot Video Editing with Agentic Reasoning$^{*}$}
\author{
    Chenyang Wu\textsuperscript{\rm 1${\dagger}$}, 
    Fuchen Long\textsuperscript{\rm 2${\dagger}$}, 
    Binyuan Huang\textsuperscript{\rm 2},
    Xinlong Sun\textsuperscript{\rm 2\ddag},
    Xi Chen\textsuperscript{\rm 2},\\
    Chun-Le Guo\textsuperscript{\rm 1},
    Chongyi Li\textsuperscript{\rm 1\S}\\
}
\affiliations{
    \textsuperscript{\rm 1}VCIP, CS, Nankai University,
    \textsuperscript{\rm 2} Smart Creation Platform Department, Online Video BU, Tencent \\ 

    chenyangwu@mail.nankai.edu.cn,\\
    \{erwinlong, shayanhuang, xinlongsun, jasonxchen\}@tencent.com,\\
    \{guochunle, lichongyi\}@nankai.edu.cn \\
}

\begin{document}
\maketitle

\begingroup
\renewcommand{\thefootnote}{$*$}
\footnotetext{Work done during the Tencent Qingyun Program internship.}
\renewcommand{\thefootnote}{$\dagger$}
\footnotetext{These authors contributed equally.}
\renewcommand{\thefootnote}{$\ddag$}
\footnotetext{Project Leader.}
\renewcommand{\thefootnote}{$\S$}
\footnotetext{Corresponding Author.}
\endgroup

\begin{abstract}
While generative AI has significantly advanced video editing, existing methods primarily focus on single-shot or short video clips. Editing long videos with multiple instructions remains a formidable challenge. Naive chunking strategies, e.g., fixed-duration segmentation, often lead to entity fragmentation, severe editing hallucinations, and disrupted temporal continuity. To bridge this gap, we introduce the Multi-Instruction Multi-Shot Long-Video Editing (MMLVE) task, which is structured around three core objectives: Cross-Shot Editing Consistency (CSEC), Multi-Instruction Decoupling (MID), and Zero-Destruction on Spatiotemporal Structure (ZDSS). To tackle these three unique challenges, we introduce an agentic editing framework that leverages the synergy of Large Language Models (LLMs) and Vision-Language Models (VLMs) to achieve shot-level video decoupling and precise instruction parsing. Furthermore, to comprehensively evaluate this task, we construct MMLVE-Bench, which is an MMLVE-focused dataset characterized by complex real-world spatiotemporal dynamics, high-density heterogeneous instructions, and sparse, random entity distributions. Three MMLVE-focused evaluation metrics are further exploited to assess the quality of the editing results. Extensive experiments demonstrate that our MMLVE-Agent outperforms existing closed-source SOTA approaches (e.g., Seedance 2.0), successfully eliminating editing hallucinations, preserving cross-shot editing consistency, and attaining seamless spatiotemporal transitions. 

\begin{links}
\link{Project Page}{https://wucy0519.github.io/MMLVE/}
\end{links}

\end{abstract}

\section{Introduction}
\label{sec:intro}

\begin{figure}[t]
    \centering
    \includegraphics[width=\linewidth]{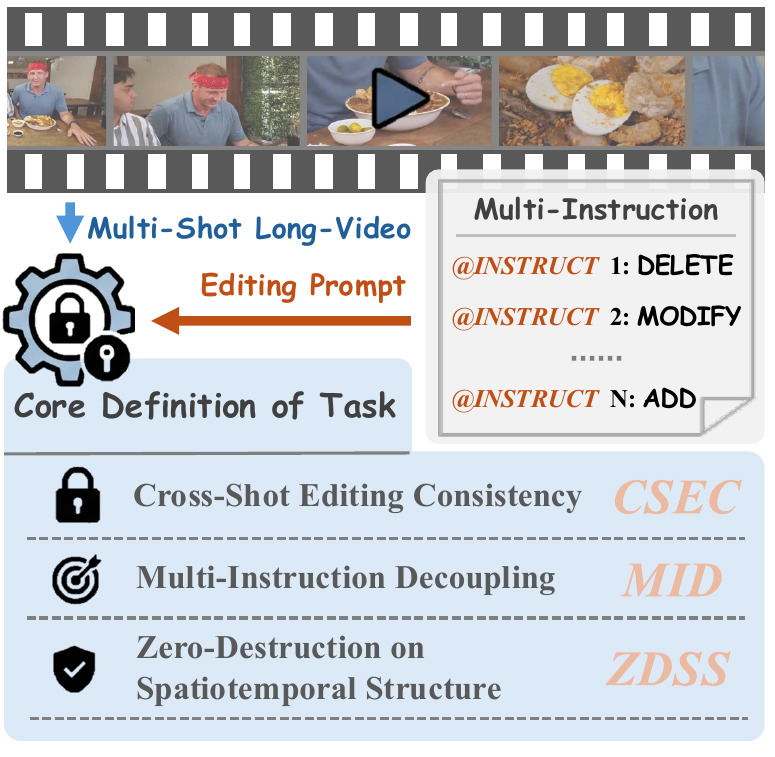}
    \vspace{-0.6cm}
    \caption{Task Definition of MMLVE. The core task entails three key constraints: 
    (1) {\textbf{CSEC}} : Cross-Shot Editing Consistency, preserving entity attributes across transitions; 
    (2) {\textbf{MID}} :  Multi-Instruction Decoupling, ensuring independent execution of diverse commands; 
    and (3) {\textbf{ZDSS}} :  Zero-Destruction on Spatiotemporal Structure, maintaining non-edited backgrounds and temporal continuity.
    }
    \vspace{-0.5cm}
    \label{fig:teaser}
\end{figure}

Generative Artificial Intelligence has catalyzed a paradigm shift in video editing, empowering users to manipulate visual content through intuitive natural language instructions~\cite{jiang2025vace,chen2025iccv,song2026soap2soap, DBLP:journals/corr/abs-2606-05833, DBLP:conf/aaai/WuFGHL26}. 
Recent advancements in diffusion models and video foundation models have demonstrated unprecedented capabilities in generating~\cite{wan2025wan,long2024vs} and editing high-fidelity videos~\cite{wu2026yose, fan2024videoagent,reco2026,coinve2026}. 
However, these successes are predominantly confined to short, single-shot video clips (typically under 15 seconds) with simple, homogeneous instructions. 
In real-world applications, videos are inherently long, characterized by complex spatiotemporal dynamics, frequent shot transitions, and sparse entity distributions~\cite{song2026soap2soap}. 
Editing such videos with multiple, heterogeneous instructions poses a formidable challenge that remains largely underexplored.

Existing long-video editing baselines~\cite{seedance2026seedance, team2023gemini} often resort to a naive fixed-duration chunking strategy (e.g., segmenting a 1-minute video into several 15-second clips) and applying all instructions to every chunk simultaneously.
This ``blind chunking'' approach leads to severe degradation in editing quality. 
First, it causes editing hallucinations due to the sparsity of entity distribution.
For instance, if a user instructs to ``add a hat to the dog'', but the dog only appears in the final 10 seconds of the video, applying this instruction to the first 15-second chunk could force the model to hallucinate an unexpected dog.
Second, it fails to maintain identity consistency across different shots, as each chunk is edited independently without a global visual anchor. 
Finally, the rigid temporal segmentation inevitably disrupts the natural spatiotemporal structure, possibly resulting in flickering and abrupt transitions at the chunk boundaries.

To systematically address these limitations, we formalize a novel and challenging task: \textbf{M}ulti-Instruction \textbf{M}ulti-Shot \textbf{L}ong \textbf{V}ideo \textbf{E}diting (\textbf{MMLVE}). 
As illustrated in Fig.~\ref{fig:teaser}, MMLVE is governed by three core objectives: 
(1) Cross-Shot Editing Consistency (CSEC), ensuring the edited entity maintains a unified appearance across diverse shot transitions; 
(2) Multi-Instruction Decoupling (MID), guaranteeing that multiple editing commands are executed independently without mutual interference or hallucination; 
and (3) Zero-Destruction on Spatiotemporal Structure (ZDSS), strictly preserving the non-edited background, original camera movements, and temporal continuity.

To conquer the unique challenges of MMLVE, we propose an agentic editing framework, termed \textbf{MMLVE-Agent}, which shifts the paradigm from ``blind chunking'' to ``Thinking on Shots''. 
The proposed framework leverages the synergy of Large Language Models (LLMs) and Vision-Language Models (VLMs). 
Initially, it performs physical shot detection and LLM-driven instruction parsing to decouple complex user commands. 
To prevent editing hallucinations caused by entity sparsity, we introduce a retrieval-based on-demand editing mechanism—editing operations are only triggered in shots where the target entity is explicitly detected by the VLM via a robust voting strategy. 
Crucially, to ensure Cross-Shot Editing Consistency (CSEC), we pioneer a \textit{Global Memory Card} mechanism. 
By extracting keyframes and generating a reference-to-edited image pair before processing the video, we provide the underlying video editor with a global visual anchor, ensuring the entity's appearance remains strictly aligned regardless of shot changes. 
Furthermore, to ensure Multi-Instruction Decoupling (MID) and Zero-Destruction on Spatiotemporal Structure (ZDSS), we incorporate a closed-loop \textit{Pos-Neg Editing Feedback (P-NEF)} mechanism. 
Through VLM-driven QA agents at both the image and video levels, the framework self-reflects on its editing results, iteratively refining corrective prompts to reduce the number of editing attempts.

Recognizing the absence of suitable benchmarks for this task, we construct \textbf{MMLVE-Bench}, a meticulously curated dataset derived from long videos.
It features complex camera movements, high-density heterogeneous instructions (e.g., ADD, DELETE, MODIFY), and random entity distributions. 
Alongside the dataset, we propose three quantitative metrics specifically designed to measure CSEC, MID, and ZDSS, providing a standardized testbed for future research.

In summary, our main contributions are threefold:
\begin{itemize}
    \item We formalize the Multi-Instruction Multi-Shot Long-Video Editing (MMLVE) task and define three core objectives (CSEC, MID, ZDSS) to address the critical flaws of existing chunking-based editing methods.
    \item We propose MMLVE-Agent, a novel framework featuring the Global Memory Card and the P-NEF closed-loop mechanism, enabling precise instruction decoupling, cross-shot consistency, and hallucination-free editing.
    \item We carefully construct MMLVE-Bench and introduce three MMLVE-focused evaluation metrics to assess the quality of the multi-shot video editing results. Extensive experiments demonstrate that our framework significantly outperforms existing state-of-the-art models.
\end{itemize}

\begin{figure*}[t!]
    \centering
    \includegraphics[width=\linewidth]{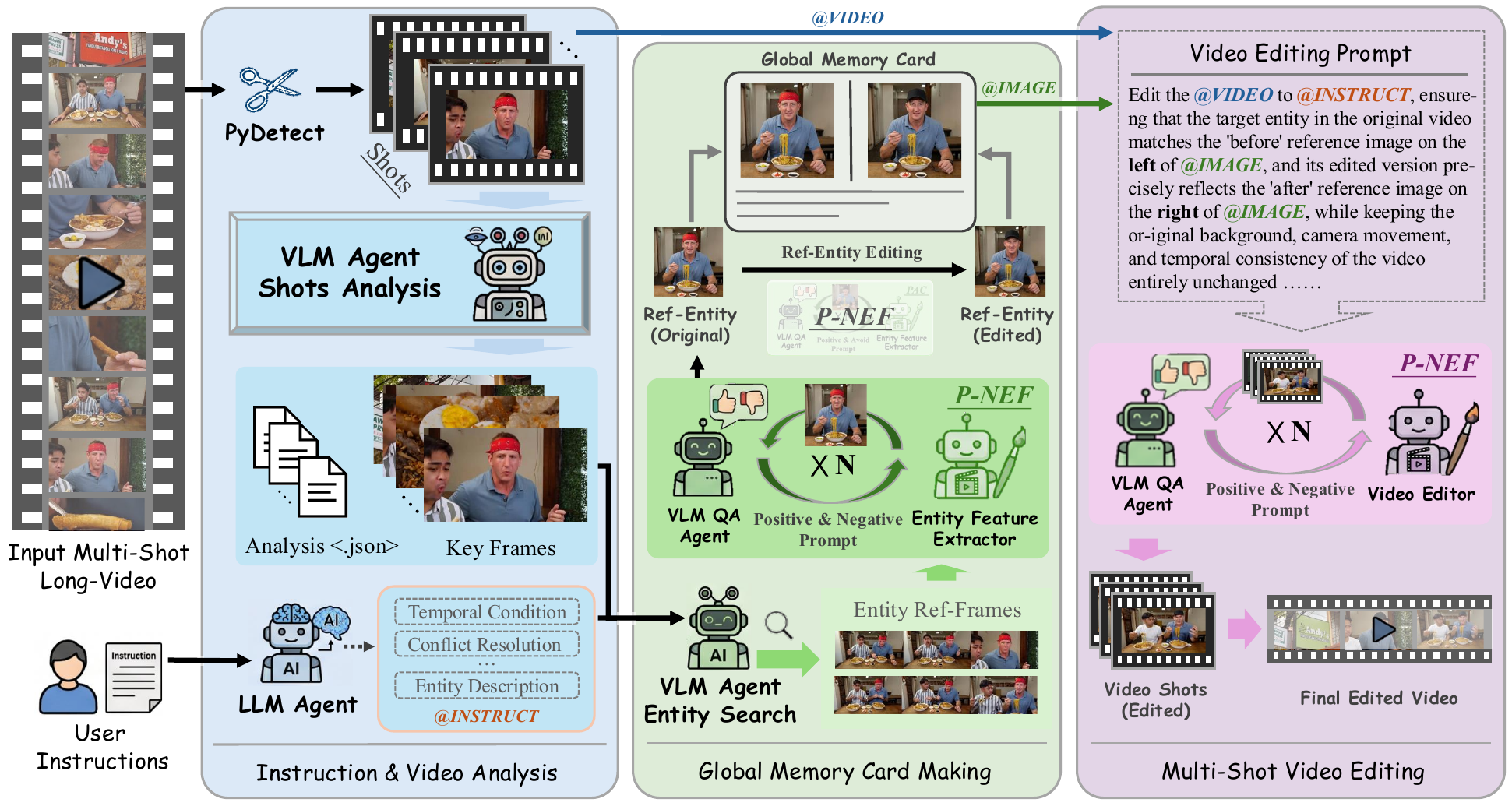}
    \vspace{-0.6cm}
    \caption{An overview of our \textbf{MMLVE-Agent} framework. 
    The pipeline comprises three core modules: 
    (1) Instruction \& Video Analysis, where the input long-video is segmented into physical shots via PyDetect. 
    Then, an LLM Agent parses and decouples complex user instructions, while a VLM Agent extracts keyframes and conducts shot-by-shot plot analysis to capture scene dynamics;
    (2) Global Memory Card Making, where target entities are retrieved via VLM voting, and the Global Memory Card, a global visual editing anchor, is generated. 
    This card is iteratively refined through an image-level Pos-Neg Editing Feedback (P-NEF) to ensure Cross-Shot Editing Consistency (CSEC); 
    and (3) Multi-Shot Video Editing, which executes retrieval-based on-demand editing. 
    A video-level P-NEF loop self-corrects the generated shots to precisely ensure Multi-Instruction Decoupling (MID) and Zero-Destruction on Spatiotemporal Structure (ZDSS) in each segmented shot before the final video combination.}
    \vspace{-0.5cm}
    \label{fig:framework}
\end{figure*}

\section{Related Work}
\label{sec:relatedwork}

\noindent \textbf{Multi-Agent System. } 
Recent advancements in LLMs and VLMs have spurred autonomous agents for complex reasoning and tool orchestration~\cite{shen2023hugginggpt, hong2024metagpt, DBLP:conf/siggrapha/ZhangXYYLXWWHZL25, DBLP:journals/corr/abs-2512-22536}. 
While VLM-based agents like UniVA~\cite{liang2025univa} excel in video understanding, deploying them for complex, generative video editing remains largely unexplored. 
Our MMLVE-Agent pioneers a heterogeneous multi-agent architecture to decouple complex editing instructions. 
Unlike standard open-loop agents, we introduce a Pos-Neg Editing Feedback (P-NEF) mechanism, enabling autonomous self-correction and monotonic improvement in visual generation.

\noindent \textbf{Long-Video Editing. }
Text-driven video editing has achieved remarkable progress via diffusion models~\cite{wu2023tune, bar2022text2live}. 
However, these methods~\cite{yu2026aurora} are primarily optimized for short clips. 
To handle longer videos, recent works employ sliding windows~\cite{li2025stable} or latent interpolation~\cite{ge2022long, team2025longcat}.
Yet, when faced with long videos containing sparse entity distributions and heterogeneous instructions, these ``blind chunking'' strategies inevitably suffer from severe editing hallucinations and temporal disruption~\cite{song2026soap2soap}. 
Our work addresses this bottleneck by shifting the paradigm to ``Thinking on Shots'', ensuring precise temporal localization and Zero-Destruction on Spatiotemporal Structure (ZDSS).

\noindent \textbf{Multi-Shot Video Editing. }
Real-world videos are inherently multi-shot, characterized by frequent camera cuts, varying background scales, and abrupt scene transitions~\cite{wang2026echoshot, DBLP:journals/corr/abs-2512-19539}. 
While single-shot video generation has been extensively studied, multi-shot scenarios introduce the daunting challenge of maintaining cross-shot consistency. 
Recent efforts in multi-shot generation, such as StoryDiffusion~\cite{zhou2024storydiffusion} and Soap2soap~\cite{song2026soap2soap}, attempt to generate consistent characters across different scenes using reference images or autoregressive conditioning. 
However, these methods~\cite{song2026soap2soap, huang2026vimax} focus primarily on generation from scratch rather than editing existing complex videos. 
Editing multi-shot videos requires not only preserving the identity of the edited entity across transitions but also executing diverse instructions independently without mutual interference.
To tackle this, we introduce the Global Memory Card mechanism, providing a unified visual anchor that guarantees Cross-Shot Editing Consistency (CSEC) and Multi-Instruction Decoupling (MID) across arbitrary physical shots.

\section{Methodology}
\label{sec:method}
In this section, we present the proposed MMLVE-Agent framework in detail.
Firstly, we formally define the \textbf{M}ulti-Instruction \textbf{M}ulti-Shot \textbf{L}ong \textbf{V}ideo \textbf{E}diting (\textbf{MMLVE}) task and its core objectives.
Subsequently, we elaborate on the overall architecture of our agentic framework, MMLVE-Agent, which is designed to address the inherent challenges of MMLVE.
Finally, we introduce the construction of the MMLVE-Bench, a benchmark for MMLVE evaluation.

\subsection{Task Definition of  MMLVE}
\label{sec:sub_task_def}
Given an input multi-shot long-video ${\mathcal{V}}$ and a complex user prompt containing a wide variety of editing instructions, ${\mathcal{I}=\{I_0, \dots, I_N\}}$, the goal of the MMLVE task is to generate an edited video ${\mathcal{V^{\dagger}}}$ that accurately reflects all instructions while maintaining the original video's structure, particularly those parts that do not require modification.
Unlike short video clips, a long-video ${\mathcal{V}}$ inherently consists of multiple non-overlapping physical shots ${\mathcal{S}=\{S_0,\dots, S_N\}}$ due to camera cuts and scene transitions.
Furthermore, the target entities $\{E_0, \dots, E_N\}$ associated with the instructions $\mathcal{I}$ often exhibit sparse and random temporal distributions (i.e., an entity $E_k$ may only appear in a specific subset of shots). 
To successfully edit such complex videos, the generated $\mathcal{V}^\dagger$ must strictly adhere to three core constraints:

\noindent \textbf{Cross-Shot Editing Consistency (CSEC):} 
When a target entity $E_k$ is edited according to instruction $I_k$, its modified appearance $E_k^\dagger$ must remain visually unified across all shots where it appears. 
%
    
\noindent \textbf{Multi-Instruction Decoupling (MID):} 
The execution of the instruction set $\mathcal{I}$ must be mutually independent. 
This requires the model to precisely map each instruction $I_k$ to its corresponding entity $E_k$ and, crucially, to its correct temporal window. 
If $E_k$ is absent in a specific shot $S_m$, the model must not hallucinate the entity or apply $I_k$ to $S_m$, thereby eliminating instruction interference and editing hallucinations.
    
\noindent \textbf{Zero-Destruction on Spatiotemporal Structure (ZDSS):} 
The editing operations must be strictly localized to the target entities. The unedited regions (e.g., backgrounds, non-target objects) and the underlying temporal dynamics (e.g., camera movements, natural motion flow) must remain identical between the original video $\mathcal{V}$ and the edited video $\mathcal{V}^\dagger$. 

By defining these three constraints, we also establish a standard for evaluating the true capabilities of video editing models in multi-shot long-form scenarios.

\subsection{Overall Architecture of MMLVE-Agent}
As shown in Fig.~\ref{fig:framework}, the MMLVE-Agent framework processes the input long-video and complex instructions through three collaborative modules: Instruction \& Video Analysis, Global Memory Card Making, and Multi-Shot Video Editing.

\noindent \textbf{Instruction \& Video Analysis. }
The first stage aims to decouple the complex spatiotemporal dynamics of the input video and the heterogeneous user instructions. 
Given a long-video $\mathcal{V}$, we employ PyDetect to perform physical shot boundary detection, segmenting the video into a sequence of independent shots $\mathcal{S}=\{S_0,\dots, S_N\}$. 
Concurrently, an LLM Agent is deployed to parse the raw user prompt. 
It resolves potential instruction conflicts, extracts temporal conditions, and decouples the prompt into a set of distinct entity descriptions and their corresponding editing instructions $\mathcal{I}$. 
Subsequently, a VLM Agent analyzes each segmented shot, extracting representative keyframes and conducting shot-by-shot plot analysis to capture the underlying scene dynamics. This dual-path analysis lays the foundation for precise temporal localization.

\begin{figure}[t]
    \centering
    \includegraphics[width=\linewidth]{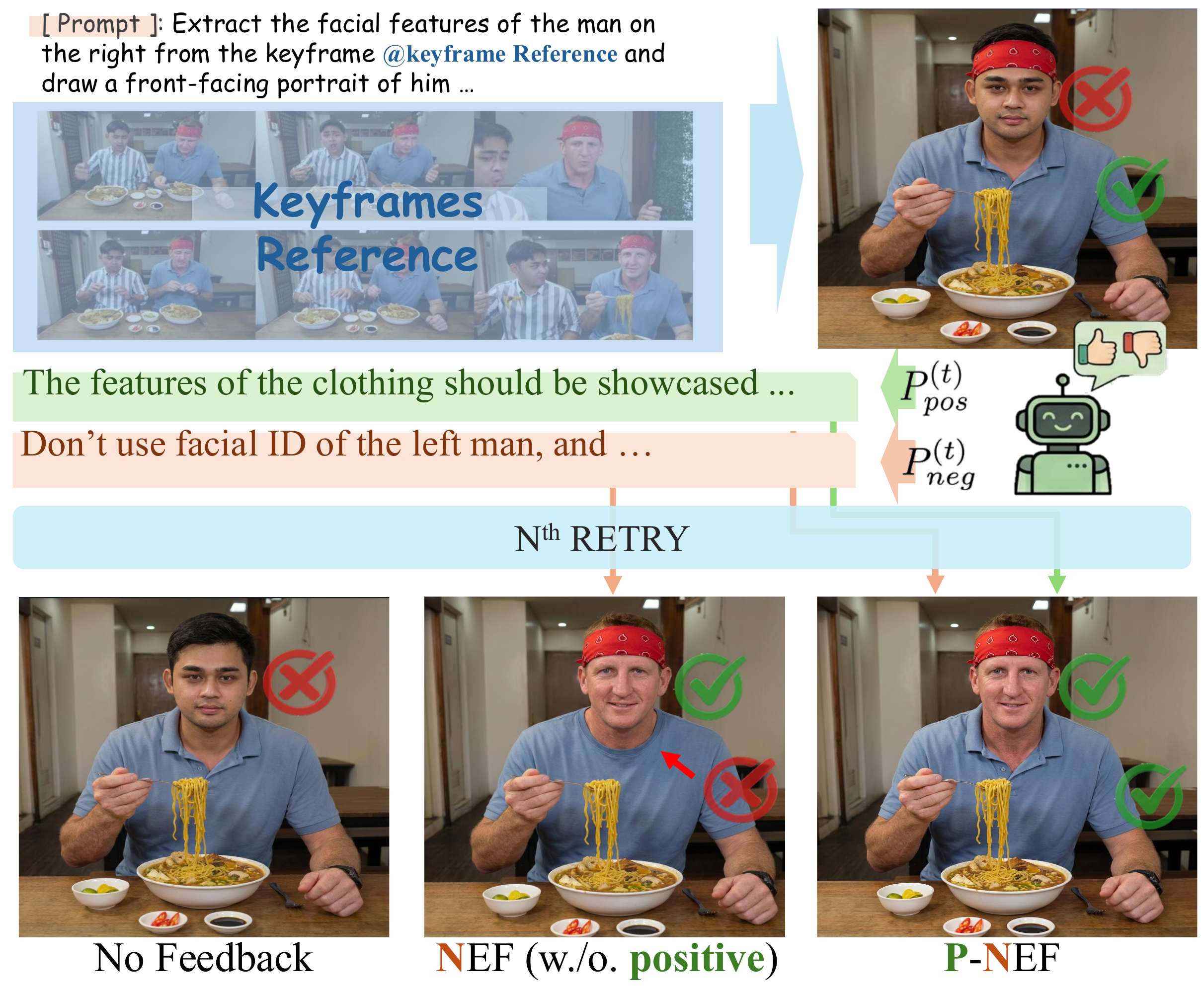}
    \vspace{-0.6cm}
    \caption{Pos-Neg Editing Feedback (P-NEF) mechanism. 
    When faced with complex instructions (grid-shape reference image or multiple instructions), the initial generation often suffers from attribute interference (e.g., incorrectly extracting the left man's facial ID). 
    To address this, the VLM evaluator generates dual feedback: a Negative Prompt ($P_{neg}$) to correct errors and a Positive Prompt ($P_{pos}$) to act as a balancer for the model’s attention. 
    Crucially, as shown in the bottom row, solely relying on the Negative Prompt (NEF w./o. positive) causes the model to focus more on ``avoid items'' in the prompt, resulting in changes to parts that do not require editing (e.g., the T-shirt's feature of the man). 
    }
    \vspace{-0.5cm}
    \label{fig:pac_show}
\end{figure}

\noindent \textbf{Global Memory Card Making. }
To achieve Cross-Shot Editing Consistency (CSEC), we propose the Global Memory Card mechanism, which serves as a global visual editing anchor for the underlying video editor. 
As shown in Fig.~\ref{fig:framework}, first, based on the extracted entity descriptions, the VLM Agent queries the keyframes of all shots to retrieve frames containing the target entity. 
The top-k (k=6) keyframes with the highest confidence scores are aggregated into a reference grid. 
Conditioned on this grid and the entity description, the generation model $\mathcal{G}$ synthesizes an initial reference image of the original entity. 

As shown in Fig.~\ref{fig:pac_show}, to ensure absolute fidelity, we introduce an image-level \textbf{Pos-Neg Editing Feedback} (P-NEF).
A VLM QA Agent acts as an evaluator $\mathcal{E}_{img}$ to verify whether the generated reference matches the keyframes. 
At $t^{th}$ attempt, the evaluator outputs a binary verification score $v^{(t)} \in \{0, 1\}$ and a feedback tuple:
\begin{equation}
v^{(t)}, \left( P^{(t)}_{pos}, P^{(t)}_{neg} \right) = \mathcal{E}_{img}(Img^{(t)}, D^{(t)}),
\label{eq:eq0}
\end{equation} 
where $D^{(t)}$ means editing prompt of the $t$ steps, and $P^{(t)}_{pos}$ is the ``Positive Prompt'' (features to retain) and $P^{(t)}_{neg}$ is the ``Negative Prompt'' (artifacts to avoid). 
The rationale for this dual-prompt strategy stems from the inherent challenges of iterative prompt refinement. 
Solely relying on a ``Negative Prompt'' forces the model to process an accumulating list of ``avoidance'' constraints, which inevitably causes attention drift.  
Specifically, the model's attention balance shifts disproportionately towards the ``avoid items'', resulting in insufficient attention being paid to the regions that were already correctly edited~\cite{chefer2023attend, hertz2022prompt}. 
Consequently, previously accurate structures or features are inadvertently altered or lost during the re-generation process. 
To counteract this, the ``Positive Prompt'' acts as a crucial semantic anchor. 
By explicitly reinforcing the successfully generated attributes, it balances the model's attention and ensures monotonic improvement across prompt refinement iterations.
If $v^{(t)} = 0$, the generation prompt is updated via concatenation ($\oplus$):
\begin{equation}
 D^{(t+1)} = D^{(t)} \oplus P^{(t)}_{pos} \oplus P^{(t)}_{neg} ,
\label{eq:eq1}
\end{equation} 
a new image $Img^{(t+1)} = \mathcal{G}(D^{(t+1)}, K)$ is generated. 
This self-correction loop repeats until $v^{(t)} = 1$ or reaches the maximum attempts $T_{max}=3$. 
If all attempts fail, the VLM selects the best candidate $Img^* = \arg\max_{t} \text{Score}(Img^{(t)})$.
Finally, the VLM extracts detailed visual features from this validated reference image to enrich the original text description. 
Using P-NEF as well, the reference image is edited according to the user instruction, yielding the Global Memory Card, which is a side-by-side visual prompt demonstrating the exact ``before-and-after'' states of the entity.

\noindent \textbf{Multi-Shot Video Editing. }
The final stage executes the actual video manipulation while strictly enforcing Multi-Instruction Decoupling (MID) and Zero-Destruction on Spatiotemporal Structure (ZDSS). 
To eliminate editing hallucinations caused by sparse entity distributions, we implement a retrieval-based on-demand editing strategy. 
For each shot $S_i$, the VLM conducts a 3-time voting mechanism using the enriched entity description and the Global Memory Card. 
The editing operation is triggered if and only if the entity receives at least 2 positive votes; otherwise, the shot is skipped and preserved in its original state.
For shots where the entity is present, the original shot, the enriched description, the instruction, and the Comparison Card are fed into the Video Editor (e.g., HappyHorse). 
To further ensure ZDSS, we employ a video-level P-NEF feedback loop. 
The VLM QA Agent extracts keyframes (determined in accordance with the \textit{Instruction\&Video Analysis} part) from the edited shot to verify two criteria: (1) successful execution of the editing task, and (2) strict preservation of non-edited regions (backgrounds and original motion). 
Similar to the image-level P-NEF, corrective prompts are generated to re-edit the shot. 
Ultimately, the successfully edited shots and the untouched skipped shots are seamlessly concatenated to form the final edited long-video $\mathcal{V}^\dagger$.

\begin{figure}[t]
    \centering
    \includegraphics[width=\linewidth]{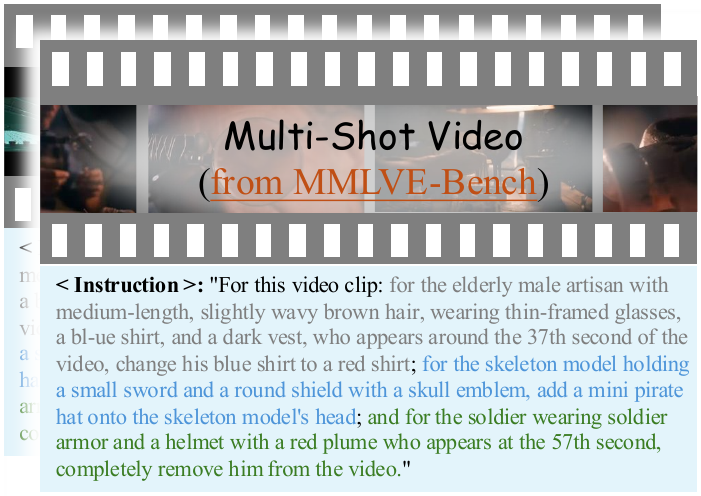}
    \vspace{-0.6cm}
    \caption{An example in MMLVE-Bench. 
    The figure illustrates a typical complex user prompt associated with a multi-shot long-video in MMLVE-Bench.
    To highlight the heterogeneous nature of the editing commands, different instruction types are color-coded: \textcolor[rgb]{ 0.5,  0.5,  0.5}{\textbf{text in grey}} denotes a MODIFY instruction, \textcolor[rgb]{ 0.38,  0.58,  0.83}{\textbf{text in blue}} indicates ADD, and \textcolor[rgb]{ 0.3,  0.5,  0.19}{\textbf{text in green}} represents DELETE.
    }
    \vspace{-0.4cm}
    \label{fig:bench_show}
\end{figure}

\subsection{MMLVE-Bench}
To comprehensively evaluate the proposed task, we construct MMLVE-Bench, a manually curated dataset derived from the open-source UniVA-Bench~\cite{liang2025univa}.
Specifically, we select 25 high-quality multi-shot long-video clips, each lasting approximately one minute, with salient editable entities.
The annotation process follows a rigorous semi-automatic pipeline: initially, a VLM is employed to detect salient entities across the video and generate candidate editing prompts, yielding about 5 distinct editing instructions per video. 
Subsequently, human annotators conduct manual verification to correct erroneous instructions and refine the textual descriptions, ensuring the high quality and logical consistency of the final prompts.

As shown in Fig.~\ref{fig:bench_show}, MMLVE-Bench is characterized by several unique features that distinguish it from existing short-video editing datasets. 
First, it features ``Complex Spatiotemporal Dynamics in Various Scenes''. 
The videos contain rich background details and complex camera movements, providing a rigorous benchmark for evaluating the ZDSS constraint. 
Second, it presents ``High-Density and High-Random Instructions''. 
The dataset encompasses a diverse range of operation types, which may target different entities within the same scene or the same entity across different scenes. 
%
%
Notably, these densely packed instructions target specific entities that appear sparsely at different timestamps (e.g., the 37th and 57th seconds), underscoring the extreme challenge of multi-instruction decoupling and temporal localization in the MMLVE task.

\begin{figure*}[t!]
    \centering
    \includegraphics[width=\linewidth]{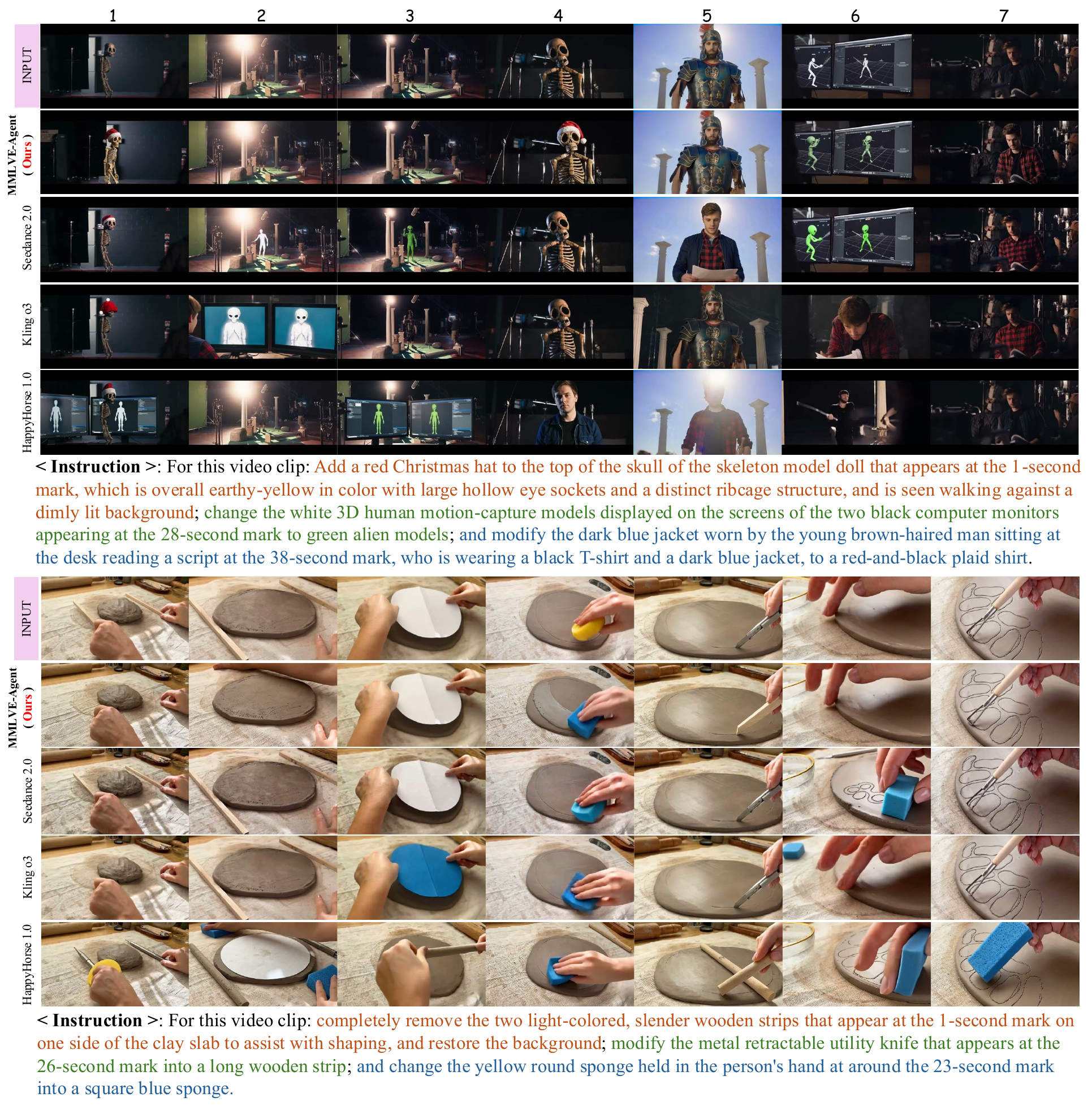}
    \vspace{-0.6cm}
    \caption{Compared with the baseline methods, only MMLVE-Agent (ours) achieves Cross-Shot Editing Consistency (CSEC), Multi-Instruction Decoupling (MID), and Zero-Destruction on Spatiotemporal Structure (ZDSS). 
    }
    \vspace{-0.4cm}
    \label{fig:exp_dx2}
\end{figure*}

\section{Experiments}
\subsection{Implementation Details}
The proposed MMLVE-Agent framework is implemented as a heterogeneous multi-agent system, orchestrating specialized models to handle distinct modalities and tasks. 
To drive the cognitive and analytical capabilities of our framework, we employ Gemini 3.5 Flash~\cite{team2023gemini} as the core Vision-Language Model (VLM). It serves as the unified backbone for the LLM Agent, the VLM Agent, and the VLM QA Evaluator, efficiently handling complex tasks ranging from instruction decoupling and keyframe retrieval to driving the Pos-Neg Editing Feedback (P-NEF) loop. 
For the visual generation tasks, specifically within the Global Memory Card Making module, we utilize the Nano Banana 2~\cite{team2023gemini} image generation model to synthesize and iteratively refine the high-fidelity reference images. 
Finally, the actual spatiotemporal video manipulation in the Multi-Shot Video Editing stage is powered by the HappyHorse video editing model, which executes the localized edits conditioned on the Global Memory Card and the decoupled prompts. 

In our experiments, the physical shot boundary detection is performed using the standard PyDetect library. 
For the P-NEF mechanism, the maximum number of self-correction iterations is empirically set to $T_{max} = 3$. 
All experiments are conducted on a MacBook Pro (M5 chip).

\subsection{Baselines}
Since MMLVE is a novel and highly challenging task, there is no existing framework explicitly designed to handle multi-instruction long-video editing.
To establish a comprehensive comparison, we select three closed-source SOTA video editing and generation foundation models as our baselines. 
Because these baseline models are inherently constrained by context length and are primarily optimized for short clips, we adapt them for the MMLVE task using the standard ``naive fixed-duration chunking'' strategy. 
Specifically, the input long-video is uniformly segmented into short clips, and the full, complex user prompt (containing all instructions) is applied to every chunk. 
The evaluated baselines include:

\noindent \textbf{Seedance 2.0}~\cite{seedance2026seedance}: A recently introduced, highly advanced video editing model known for its high-fidelity visual manipulation and structural preservation capabilities in short-video scenarios.

\noindent \textbf{Kling o3}: A cutting-edge video foundation model renowned for its robust spatiotemporal generation and dynamic consistency. We utilize its V2V editing capabilities for comparison.

\noindent \textbf{HappyHorse 1.0}: This is the foundational video editing model utilized within our own MMLVE-Agent framework. 
Evaluating it as a standalone baseline is crucial, as it directly demonstrates the performance gains and the necessity of our proposed agentic workflow (i.e., instruction decoupling, Global Memory Card, and P-NEF mechanism) over direct, naive inference.

\subsection{Qualitative Evaluation}
Fig.~\ref{fig:exp_dx2} presents a visual comparison between our MMLVE-Agent and the baseline models on complex multi-shot long-videos. 
The qualitative results clearly demonstrate the critical flaws of existing naive chunking strategies and highlight the advantage of MMLVE-Agent across three key dimensions:

\noindent \textbf{CSEC Analysis: } 
Even when baseline models attempt to execute an editing instruction, they suffer from ``amnesia'' across different shots, failing to maintain the entity's visual identity. 
In the first example of Fig.~\ref{fig:exp_dx2}, the instruction requires adding a ``red Christmas hat'' to the skeleton. 
However, both Seedance 2.0 and Kling o3 fail to maintain this attribute, forgetting to generate the hat in subsequent shots (4th frame). 
Worse still, HappyHorse 1.0 completely morphs the skeleton and the armored knight into entirely different target entities from the prompt (4th and 5th frames). 
Notably, benefiting from the Global Memory Card mechanism, MMLVE-Agent establishes a robust global visual anchor, ensuring that the edited entities maintain strict CSEC without identity degradation or flickering.

\noindent \textbf{MID Analysis: } 
Applying a complex, multi-entity prompt to all video chunks simultaneously inevitably leads to severe instruction interference. 
In the first example, the instruction to change the monitor models to ``green aliens'' leaks into unedited shots: 
Seedance 2.0 erroneously morphs an unrelated person into a green alien (3rd frame). 
Similarly, in the second example, the instruction to create a ``blue sponge'' causes severe attribute bleeding in Kling o3, which incorrectly dyes the white paper blue (3rd frame).
HappyHorse 1.0 exhibits extreme hallucinations, randomly inserting unrequested yellow and blue sponges across multiple frames (1st and 7th frames). 
%

\noindent \textbf{ZDSS Analysis: } 
A glaring issue with direct inference models is the severe disruption of the original video's background and temporal structure. 
Spatially, Kling o3 and HappyHorse 1.0 forcefully hallucinate two computer monitors into completely unrelated scenes (1st - 3rd frames in the first example), destroying the original background. 
Temporally, the blind chunking strategy causes catastrophic frame misalignment. 
In the first example (6th frame), both Kling o3 and HappyHorse 1.0 arbitrarily delete the subsequent narrative shots and replace them with earlier shots, completely scrambling the chronological order. 
Our framework, by strictly operating on parsed physical shots and preserving unedited frames, successfully satisfies the ZDSS constraint.

Even though SOTA baselines suffer from instruction confusion, temporal scrambling, and identity loss, MMLVE-Agent still consistently delivers high-fidelity, hallucination-free, and spatiotemporally coherent long-video edits.

\begin{table}[h]
  \centering
    \begin{tabular}{ccccc}
    \toprule
    Method & CSEC $\uparrow$ & MID $\uparrow$ & ZDSS $\uparrow$ & Avg. $\uparrow$  \\
    \midrule
    \makecell[c]{Seedance 2.0 $\dagger$} & 77.58  & 78.58  & \textbf{82.25 } & 79.47\\
    \makecell[c]{Kling o3 $\dagger$} & 70.00  & 72.57  & 69.13  & 70.57 \\
    \makecell[c]{HappyHorse 1.0} & 74.68  & 67.28  & 66.88  & 69.61  \\
    \rowcolor[rgb]{ .9, .9, 0.9} \makecell[c]{MMLVE-Agent \\ ( ours )} &  \textbf{84.80 } & \textbf{79.04 } & 81.68  & \textbf{81.84 } \\
    \bottomrule
    \end{tabular}%
    \vspace{-0.1cm}
    \caption{Quantitative evaluation results on MMLVE-Bench.
    \textbf{BOLD}: best performance.
    $\dagger$ means there are 1-2 scenarios in which this method rejects processing due to its safety mechanism (are ignored in the mean calculation of the metrics).}
    \vspace{-0.2cm}
  \label{tab:addlabel}%
\end{table}%

\subsection{Quantitative Evaluation}
To comprehensively quantify the performance of the MMLVE task, we evaluate the generated videos across our three proposed core indicators: Cross-Shot Editing Consistency (CSEC), Multi-Instruction Decoupling (MID), and Zero-Destruction on Spatiotemporal Structure (ZDSS). 
To ensure a fine-grained and rigorous assessment, each main indicator is further decomposed into \textbf{five specific sub-dimensions} (e.g., identity preservation, hallucination suppression, chronological shot alignment). Each sub-dimension is strictly scored on a scale of 0 to 20, yielding a maximum possible score of 100 points per indicator. 
We avoid adopting traditional frame-level metrics like the CLIP~\cite{radford2021learning} score, as they inherently lack the reasoning capacity to comprehend complex multi-instruction decoupling and long-term spatiotemporal consistency. 
Instead, we adopt VLM (i.e., Gemini 3.5 Flash) as the video editing judger. 
Due to space constraints, the detailed definitions of the \textbf{15 sub-dimensions}, the comprehensive evaluation protocols, and \textbf{scores (including sub-dimensions) for each method in each scene of the MMLVE-Bench} are provided in supplementary materials.

As shown in Tab.~\ref{tab:addlabel}, our MMLVE-Agent achieves the highest average score (81.84), demonstrating its superior capability in handling complex long-video editing tasks. 
Specifically, our framework significantly outperforms all baselines in CSEC (84.80) and MID (79.04).
Notably, while Kling o3 and HappyHorse 1.0 suffer from severe spatiotemporal destruction (low ZDSS) due to aggressive but erroneous editing, Seedance 2.0 adopts a conservative strategy, which ignores instructions in complex scenes to avoid errors. 
Although this conservative approach artificially inflates its ZDSS score (82.25), it leads to severe missed edits (lower CSEC and MID). 
In contrast, MMLVE-Agent robustly executes all edits, making the negligible ZDSS drop a highly acceptable trade-off for its comprehensive superiority.

\begin{figure}[t]
    \centering
    \includegraphics[width=\linewidth]{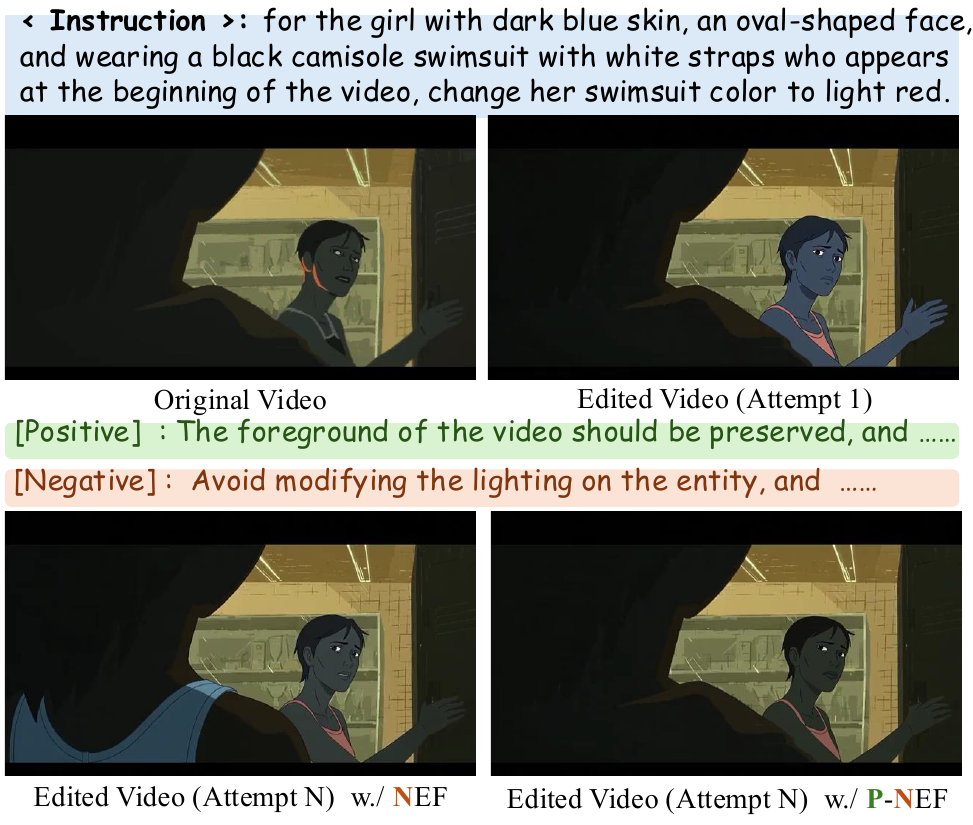}
    \vspace{-0.6cm}
    \caption{Visual Comparison between P-NEF and NEF.
    }
    \vspace{-0.4cm}
    \label{fig:ablation}
\end{figure}

\subsection{Ablation Study on P-NEF}
As shown in Fig.~\ref{fig:ablation}, we visually ablate the impact of the proposed Pos-Neg Editing Feedback (P-NEF) mechanism.
Solely relying on Negative Editing Feedback (NEF) forces the model to process an increasing number of ``avoidance'' constraints across attempts. 
This accumulation inevitably diverts the model's cross-attention, leading to a severe ``catastrophic forgetting'' effect—previously correct edits are inadvertently altered or lost. 
Consequently, this introduces new structural errors and significantly increases the required number of trial-and-error iterations. 
Conversely, the full P-NEF mechanism introduces a Positive Prompt that acts as a crucial semantic anchor. 
By explicitly reinforcing the successfully generated features, P-NEF effectively balances the model's attention, ensures monotonic improvement, and reduces the overall attempts needed to achieve a robust edit.

\section{Conclusions}
In this paper, we tackle the underexplored challenge of editing real-world, multi-shot long-video driven by complex instructions. 
We formalize the Multi-Instruction Multi-Shot Long-Video Editing (MMLVE) task, governed by three core constraints: CSEC, MID, and ZDSS.
Furthermore, we propose MMLVE-Agent and construct MMLVE-Bench, alongside tailored MMLVE-focused evaluation metrics. 

\bibliography{aaai2027}

\clearpage
\appendix

\begin{figure*}[h]
    \centering
    \includegraphics[width=\linewidth]{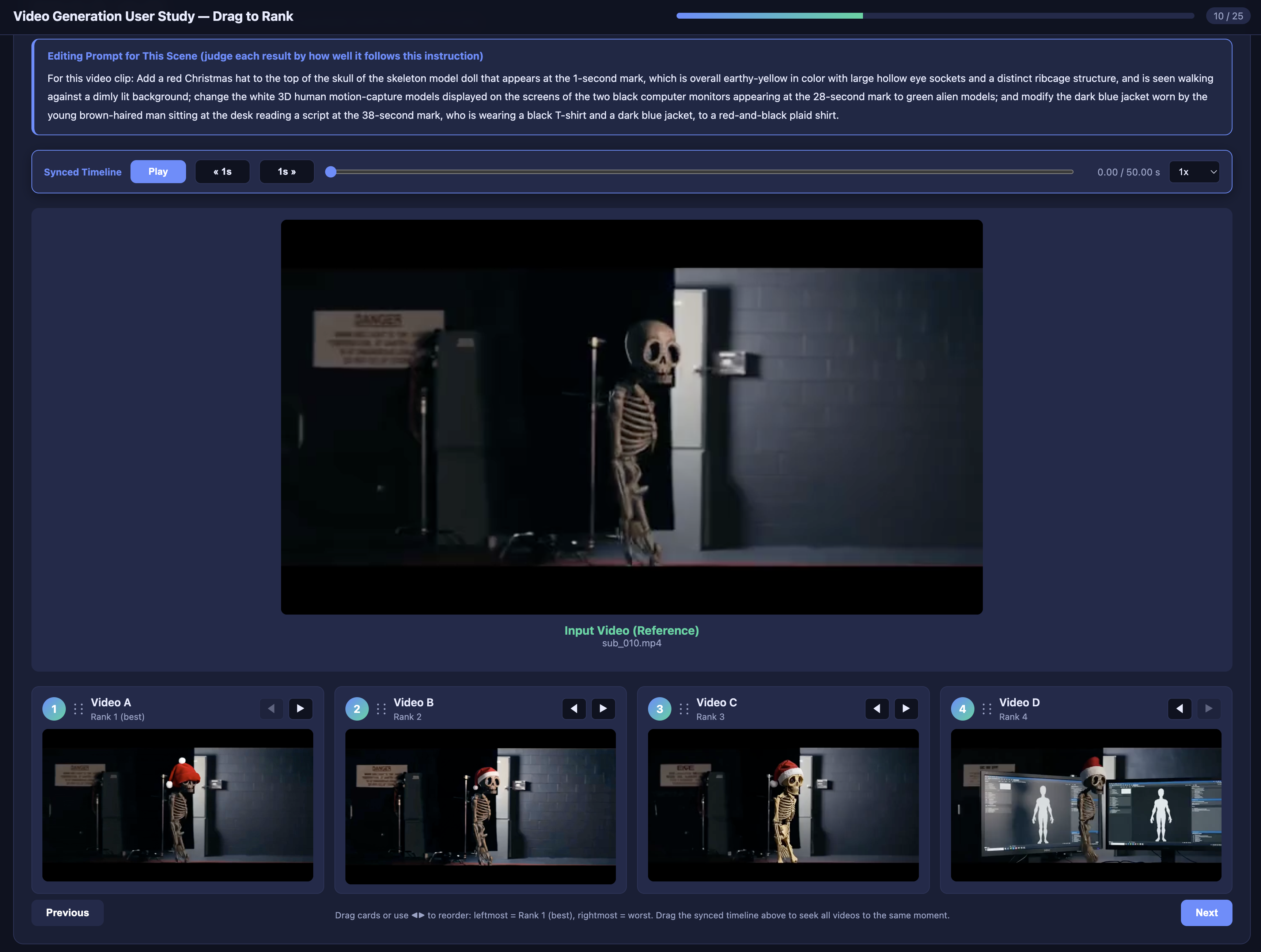}
    \caption{HTML Interface for User Study.}
    \label{fig:user_st}
\end{figure*}

\section{User Study}
\label{sec:user_study}
As shown in Fig.~\ref{fig:user_st}, to complement our automated VLM-based evaluation and assess the perceptual quality of the generated videos from a human perspective, we conducted a rigorous User Study. 
Given the inherent complexity of the MMLVE task—characterized by minute-long videos and dense, heterogeneous instructions—evaluating these results imposes a significant cognitive load. 
Therefore, we designed a specialized evaluation protocol driven by expert reviewers.

\noindent \textbf{Custom-Built Evaluation Platform.} 
As shown in Fig.~\ref{fig:user_st}, we developed a dedicated web-based evaluation interface to ensure a fair and meticulous comparison. The workflow is structured as follows:
\begin{itemize}
    \item \textbf{Initialization:} Upon accessing the platform, evaluators first read the detailed assessment guidelines, input their evaluator ID, and enter the evaluation workspace.
    \item \textbf{Blind Testing Interface:} For each scene, the UI displays the current progress, the complex editing prompt, and the original input video. The editing results from our MMLVE-Agent and the three baselines are presented side-by-side and strictly anonymized as ``Video A'', ``Video B'', ``Video C'', and ``Video D'' in a randomized order to prevent any subjective bias.
    \item \textbf{Synchronized Playback Controls:} To facilitate fine-grained spatiotemporal comparison, the platform features a unified timeline. Evaluators can synchronously play, pause (via UI buttons or the Spacebar), or scrub through all five videos (input + 4 results) simultaneously. Additionally, adjustable playback speeds are provided to allow experts to carefully inspect minute details, such as transient flickering or subtle attribute bleeding.
\end{itemize}

\noindent \textbf{Evaluation Criteria and Ranking Mechanism.} 
Evaluators are instructed to comprehensively judge the videos based on several core dimensions: (1) \textit{Execution Accuracy} (whether all instructions were correctly applied to the right entities); (2) \textit{Visual Quality \& Naturalness}; (3) \textit{Temporal Consistency}; and (4) \textit{Artifact Reduction} (absence of flickering, structural deformation, or hallucinations). 
Based on these criteria, evaluators rank the four anonymized videos from best (1st, placed on the far left) to worst (4th, placed on the far right) using an intuitive drag-and-drop mechanism or directional arrow buttons. All ranking results are automatically logged into a CSV database for subsequent statistical aggregation.

\noindent \textbf{Participants and Protocol.} 
We recruited 9 expert evaluators with extensive experience in video generation and visual content assessment. 
Because the multi-shot long videos contain rich spatiotemporal dynamics and require intense concentration to verify multiple decoupled instructions, we randomly assigned 5 distinct cases to each expert. 
This carefully controlled workload ensures that the evaluators maintain high focus and provide highly reliable, meticulous rankings for each complex scene.

\begin{table}[t]
\centering
\small
\setlength{\tabcolsep}{3.5pt}
\resizebox{0.47\textwidth}{!}{
\begin{tabular}{lccccccc}
\toprule
\multirow{2}{*}{Method} & \multicolumn{4}{c}{Rank Distribution (\%)} & Top-1 & Pairwise & Avg. Rank \\
\cmidrule(lr){2-5}
 & 1st & 2nd & 3rd & 4th & (\%) $\uparrow$ & Win (\%) $\uparrow$ & $\downarrow$ \\
\midrule
Seedance 2.0${^*}$ & 4.7 & 48.8 & 20.9 & 25.6 & 4.7 & 42.4 & 2.67 \\
Kling o3${^*}$ & 17.1 & 22.0 & 41.5 & 19.5 & 17.1 & 44.6 & 2.63 \\
HappyHorse 1.0${^*}$ & 4.4 & 15.6 & 40.0 & 40.0 & 4.4 & 24.8 & 3.16 \\
\textbf{Ours} & \textbf{75.6} & \textbf{17.8} & \textbf{2.2} & \textbf{4.4} & \textbf{75.6} & \textbf{87.6} & \textbf{1.36} \\
\bottomrule
\end{tabular}}
\caption{User study results. ``Pairwise Win'' is the head-to-head win rate against all other methods within the same ranking. $^{*}$ denotes $p<0.001$ (two-sided Wilcoxon signed-rank test against ours on paired normalized preference scores).}
\label{tab:user_study}
\end{table}

\noindent \textbf{Results and Analysis.} 
Table~\ref{tab:user_study} summarizes the 45 collected rankings. 
MMLVE-Agent is ranked first in 75.6\% of all cases and within the top two in 93.3\%, achieving the best average rank of 1.36 versus 2.63--3.16 for the baselines. 
In head-to-head comparisons it is preferred over Seedance 2.0, Kling O3 and HappyHorse in 88.4\%, 80.5\% and 93.3\% of the co-rated cases respectively; all three margins are statistically significant under a two-sided Wilcoxon signed-rank test ($p < 0.001$, Bonferroni-corrected). 
Notably, although Seedance 2.0 and Kling O3 attain nearly identical average ranks (2.67 vs.\ 2.63), their rank distributions differ substantially: Seedance 2.0 is rarely the best (4.7\% 1st) but frequently second (48.8\%), whereas Kling O3 is far more polarized (17.1\% 1st but 19.5\% last), indicating that end-to-end generators handle heterogeneous instruction sets inconsistently. 

\section{All Cases shown in MMLVE-Bench}
\label{sec:all_case}
\begin{figure*}[t]
    \centering
    \includegraphics[width=\linewidth]{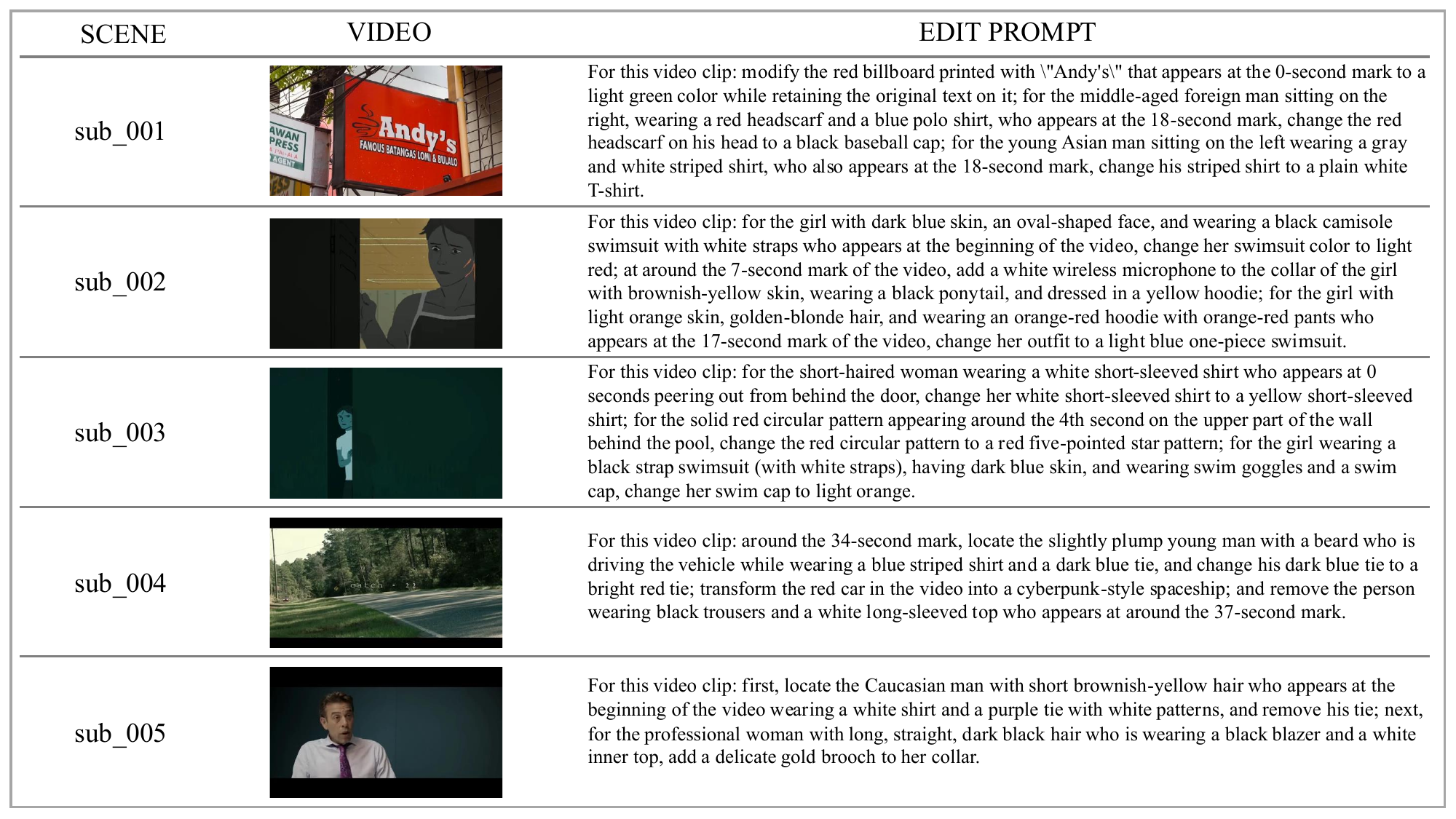}
    \caption{The cases (sub\_001 - sub\_005) in MMLVE-Bench.}
    \label{fig:dataset_show1}
\end{figure*}

\begin{figure*}[t]
    \centering
    \includegraphics[width=\linewidth]{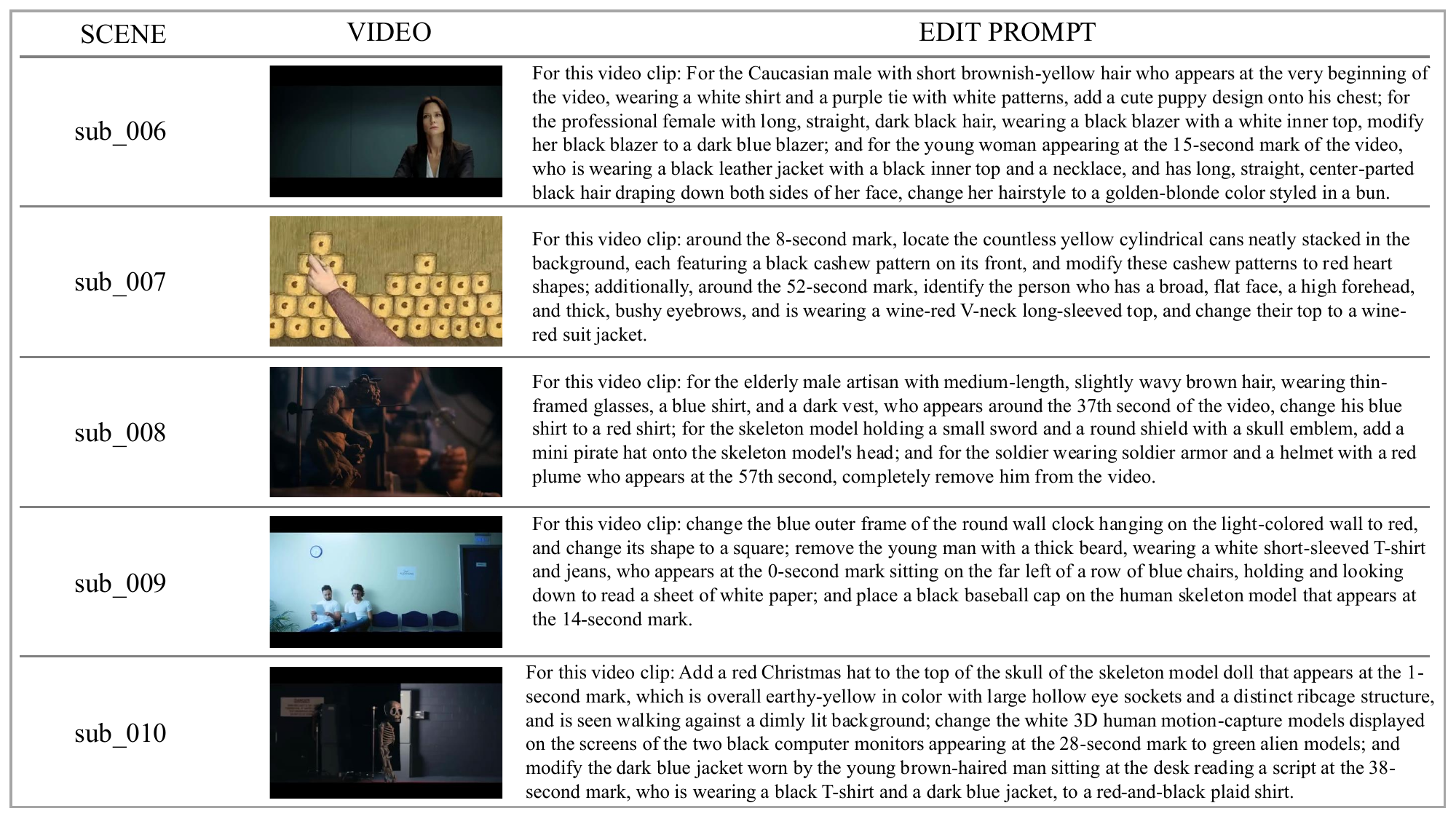}
    \caption{The cases (sub\_006 - sub\_010) in MMLVE-Bench.}
    \label{fig:dataset_show2}
\end{figure*}

\begin{figure*}[t]
    \centering
    \includegraphics[width=\linewidth]{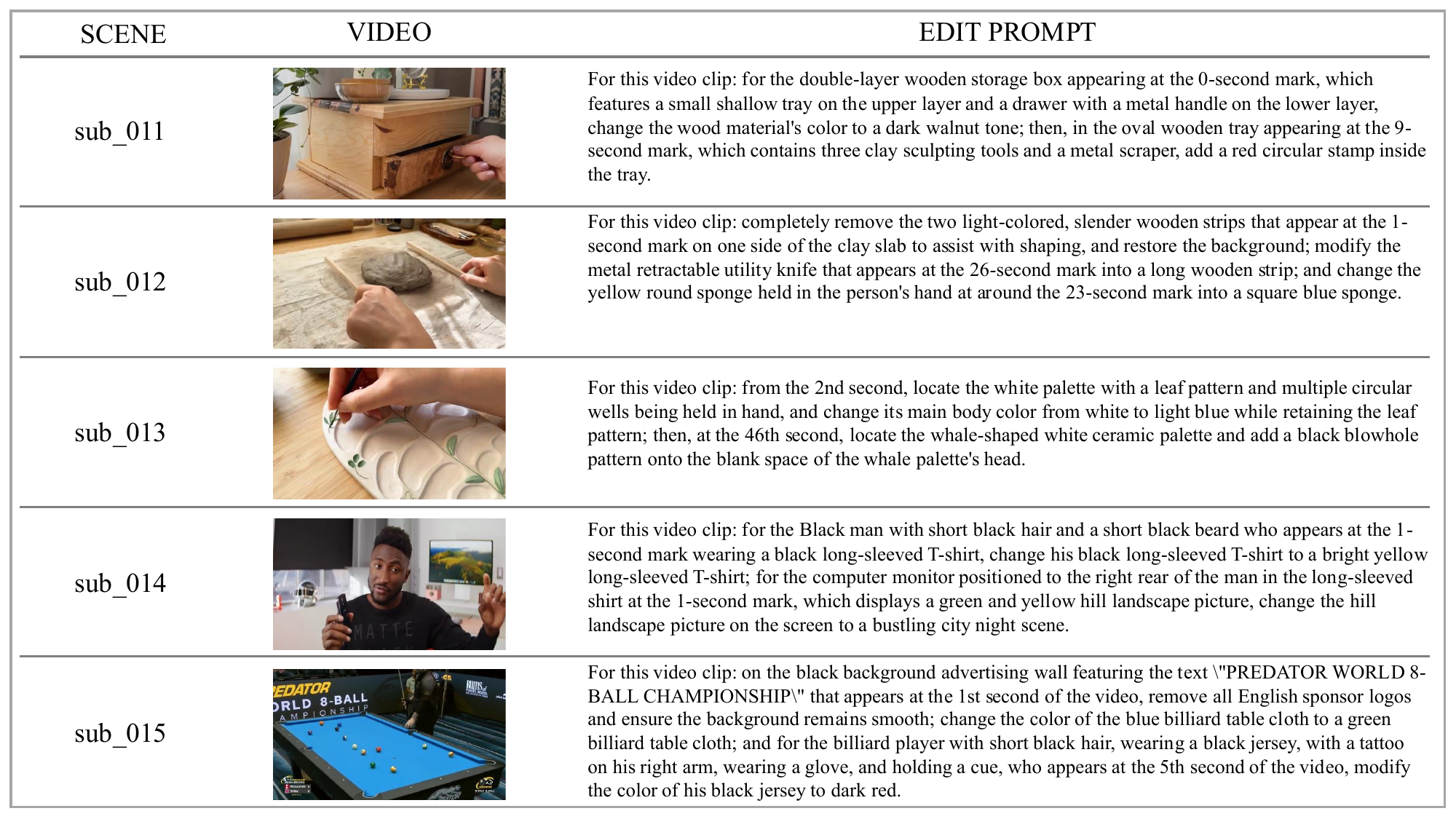}
    \caption{The cases (sub\_011 - sub\_015) in MMLVE-Bench.}
    \label{fig:dataset_show3}
\end{figure*}

\begin{figure*}[t]
    \centering
    \includegraphics[width=\linewidth]{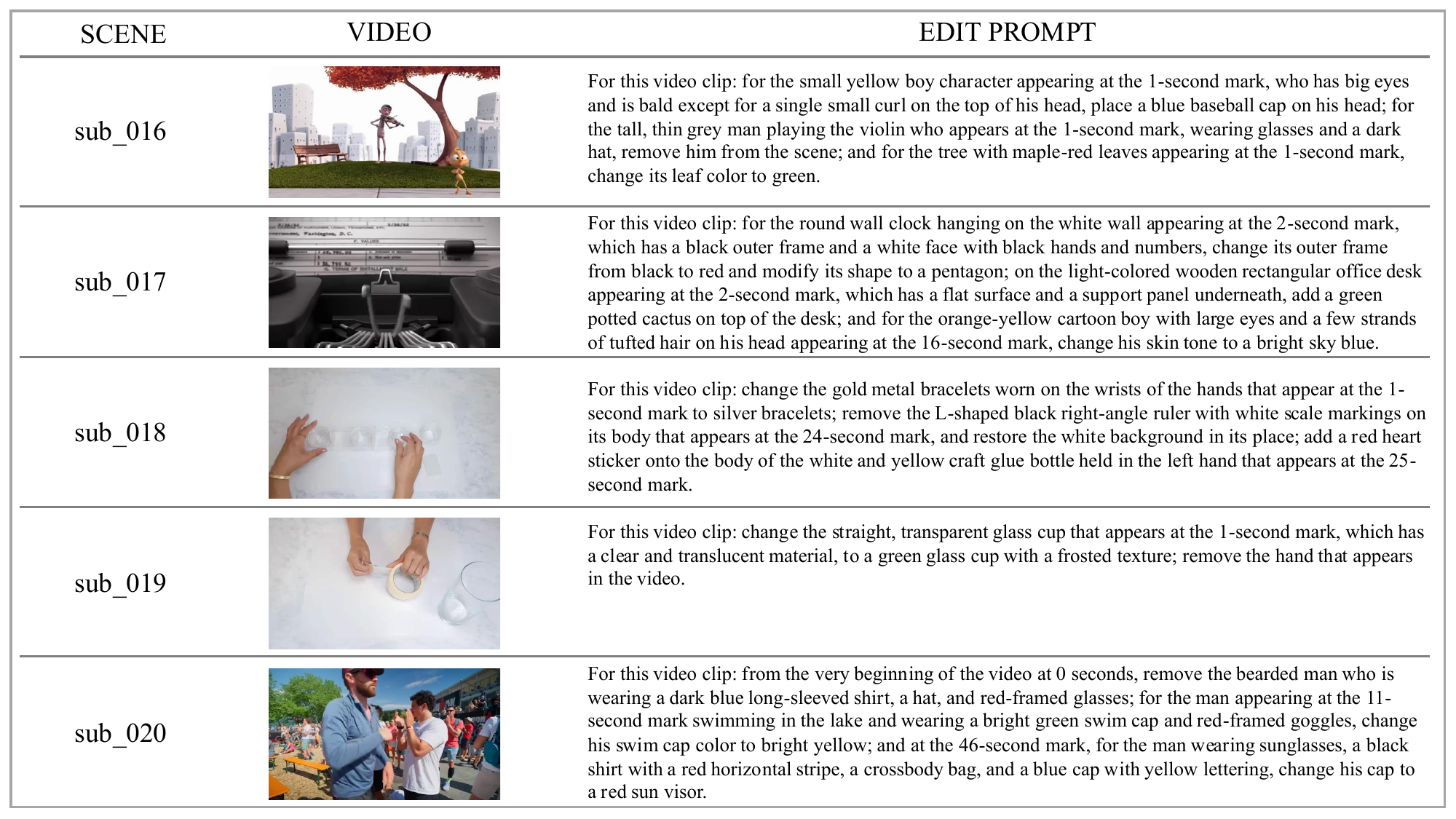}
    \caption{The cases (sub\_016 - sub\_020) in MMLVE-Bench.}
    \label{fig:dataset_show4}
\end{figure*}

\begin{figure*}[t]
    \centering
    \includegraphics[width=\linewidth]{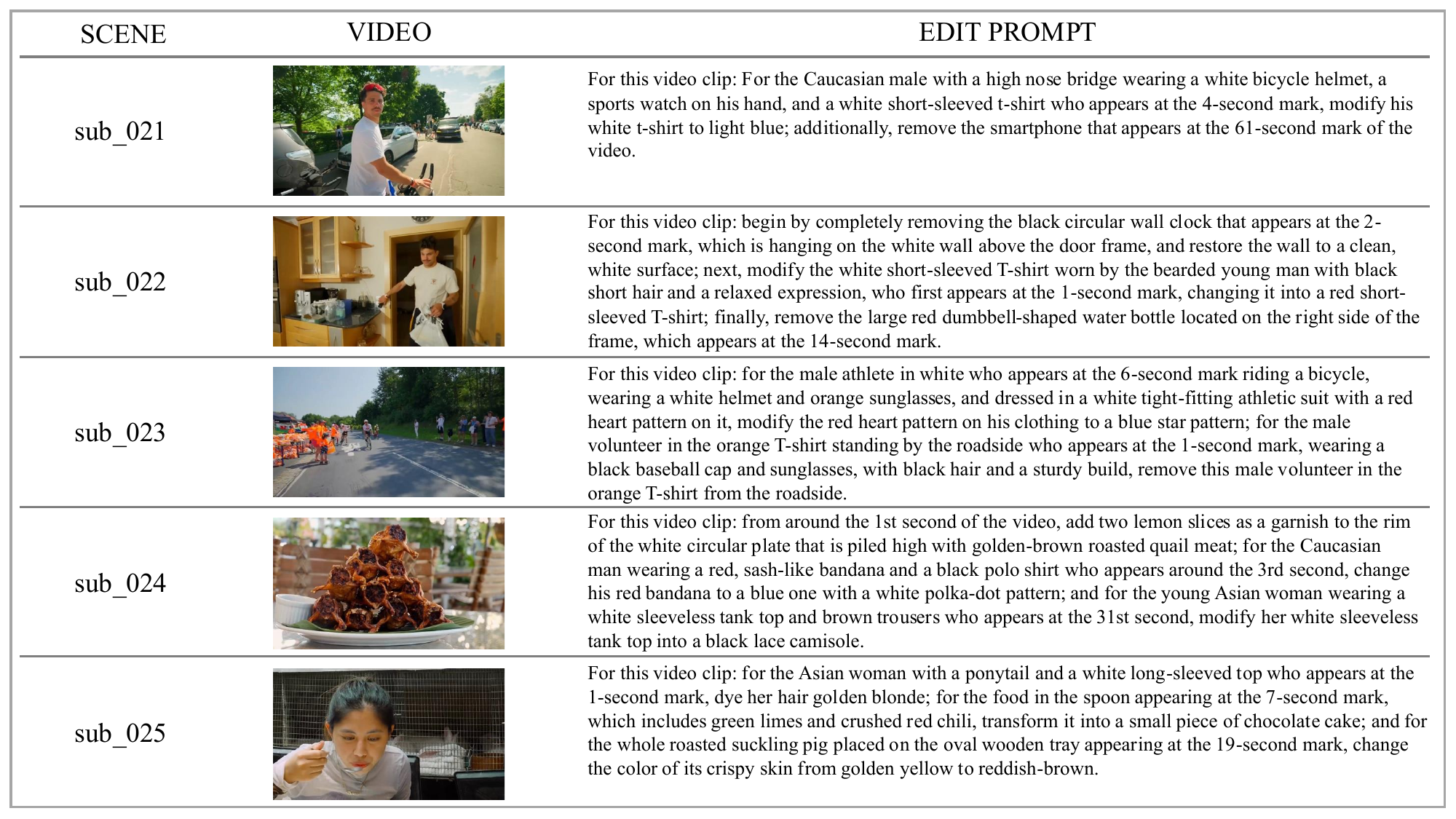}
    \caption{The cases (sub\_021 - sub\_025) in MMLVE-Bench.}
    \label{fig:dataset_show5}
\end{figure*}

To provide a comprehensive understanding of the complexity and diversity of our proposed dataset, we present all 25 curated cases from the MMLVE-Bench. 
As shown in Fig.~\ref{fig:dataset_show1} to Fig.~\ref{fig:dataset_show5}, each case consists of a representative frame from the original multi-shot long video alongside its corresponding complex editing prompt. 
These prompts feature high-density, heterogeneous instructions (e.g., ADD, MODIFY, DELETE) targeting multiple entities that appear sparsely across different physical shots. 
This exhaustive showcase highlights the extreme challenge of the MMLVE task, particularly in terms of multi-instruction decoupling and long-term spatiotemporal localization.

\begin{figure*}[t]
    \centering
    \includegraphics[width=\linewidth]{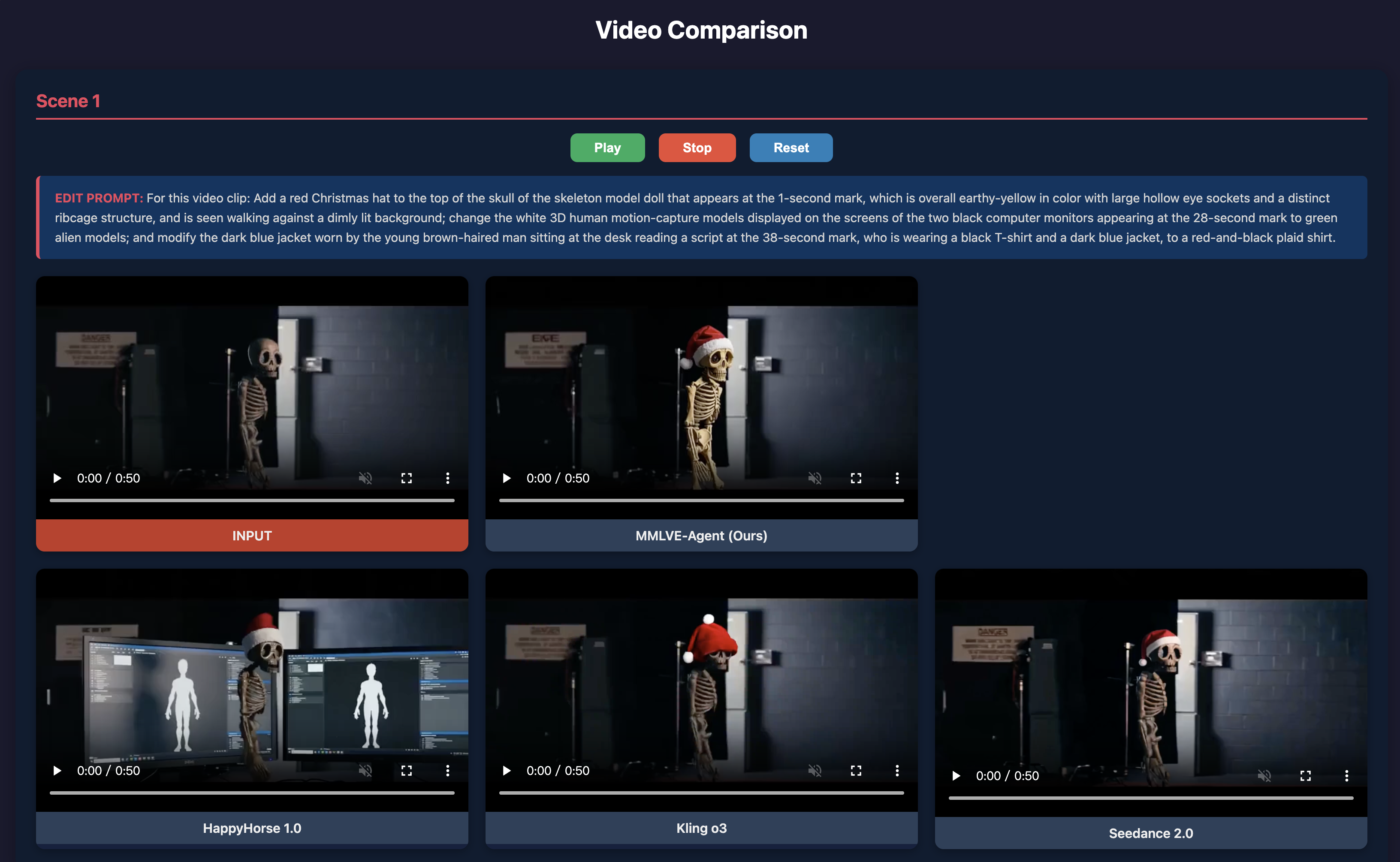}
    \caption{Video Comparison in \textcolor[rgb]{ 0.5,  0.15,  0.11}{\textbf{Media Supplement}} part.}
    \label{fig:comp_visual}
\end{figure*}

\section{Global Memory Card Show}
\label{sec:global_mem_card}
\begin{figure*}[t]
    \centering
    \includegraphics[width=\linewidth]{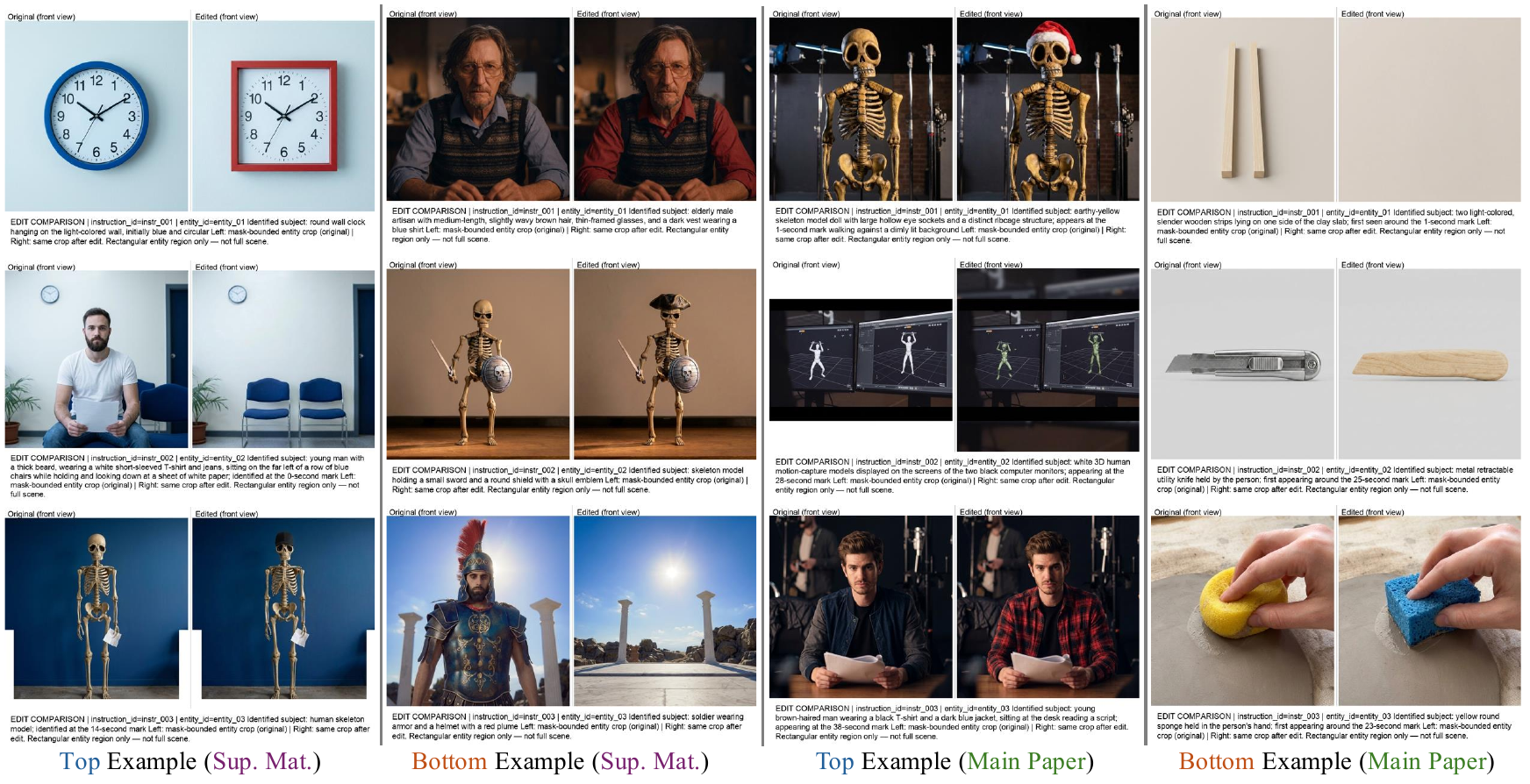}
    \caption{Global Memory Card Show.}
    \label{fig:global_mem_sup}
\end{figure*}

To further illustrate the effectiveness of our proposed Global Memory Card mechanism, we provide additional visual examples. 
As shown in Fig.~\ref{fig:global_mem_sup}, the Global Memory Card acts as a robust global visual anchor, explicitly demonstrating the exact ``before-and-after'' states of the target entities. 
By conditioning the underlying video editor on this side-by-side reference card, our MMLVE-Agent successfully maintains strict Cross-Shot Editing Consistency (CSEC) and prevents attribute interference, even in scenes with drastic camera movements, varying scales, and complex backgrounds.

\section{VLM-based Judge Evaluation Metrics}
\label{sec:vlm_judge}

Traditional frame-level metrics (e.g., CLIP score, PSNR) are inherently inadequate for evaluating long videos with complex, multi-entity instructions, as they lack the reasoning capacity to assess instruction decoupling and long-term spatiotemporal consistency. To address this, we design a robust, automated VLM-based Judge evaluation pipeline to quantitatively assess the generated videos across our three core dimensions (CSEC, MID, and ZDSS). 

\noindent \textbf{Evaluation Pipeline.} 
To ensure a strict and fair comparison, our automated evaluation script processes the results of all baseline methods through the following steps:
\begin{itemize}
    \item \textbf{Spatiotemporal Keyframe Alignment:} Instead of feeding the entire long video into the VLM, we first segment the original video into 15-second intervals. A VLM analyzes each segment to record the absolute timestamps of critical plot transitions. Based on these JSON-formatted timestamps, we extract the exact corresponding keyframes from both the original video and the edited videos of all evaluated methods. Missing generation cases from certain baselines are automatically recorded and skipped.
    \item \textbf{Multimodal Prompting:} For each scene, the original keyframes, the edited keyframes, and the complex editing prompt are simultaneously fed into the VLM judge.
    \item \textbf{Fine-Grained Scoring \& Aggregation:} The VLM evaluates the editing quality based on 15 meticulously designed sub-dimensions. The detailed scores for each sub-dimension and the main dimensions are automatically exported to a CSV file for comprehensive statistical analysis.
\end{itemize}

\noindent \textbf{Metric Definitions and Scoring Criteria.} 
Each of the three main indicators (CSEC, MID, ZDSS) is decomposed into five specific sub-dimensions. The VLM judge strictly assigns a score of 0, 10, or 20 to each sub-dimension based on predefined criteria (yielding a maximum of 100 points per main indicator):

\textbf{1. Cross-Shot Editing Consistency (CSEC):} Evaluates the visual unity of the edited entity across different physical shots (varying angles, scales, and lighting). The sub-dimensions include:
\textit{(1.1) Identity Preservation} (20=perfect identity, 0=severe deformation/amnesia); 
\textit{(1.2) Attribute Consistency} (20=attributes exist in all shots, 0=attributes lost); 
\textit{(1.3) Color \& Texture Stability}; 
\textit{(1.4) Structural \& Geometric Alignment}; 
\textit{(1.5) Lighting \& Shadow Harmony}.

\textbf{2. Multi-Instruction Decoupling (MID):} Assesses the model's ability to execute dense instructions precisely without mutual interference. The sub-dimensions include:
\textit{(2.1) Target Execution Accuracy} (20=perfect execution, 0=completely failed); 
\textit{(2.2) Distractor Isolation} (20=zero attribute bleeding, 0=severe color/texture leakage); 
\textit{(2.3) Hallucination Suppression} (20=no hallucinated entities, 0=clear generation of unrequested objects); 
\textit{(2.4) Entity Localization Precision}; 
\textit{(2.5) Semantic Conflict Resolution}.

\textbf{3. Zero-Destruction on Spatiotemporal Structure (ZDSS):} Strictly inspects the preservation of non-edited regions and chronological order. The sub-dimensions include:
\textit{(3.1) Static Background Fidelity} (20=pixel-level preservation, 0=background destroyed); 
\textit{(3.2) Non-Target Object Preservation}; 
\textit{(3.3) Chronological Shot Alignment} (20=1:1 strict alignment with original shot order, 0=severe temporal scrambling or arbitrary deletion); 
\textit{(3.4) Intra-Shot Motion Consistency}; 
\textit{(3.5) Artifact \& Flicker Reduction}.

By employing this fine-grained, 15-dimension evaluation matrix, our benchmark provides a comprehensive and interpretable quantitative assessment of multi-shot long-video editing capabilities.

\subsection{Evaluation results for each method}
\label{sec:sub_res_eval}
Due to space constraints in the main paper, the fine-grained quantitative results were aggregated. Here, we provide the exhaustive, scene-by-scene evaluation scores for all methods across the 15 sub-dimensions. 
As shown in Tab.~\ref{tab:addlabel_1} and Tab.~\ref{tab:addlabel_2}, our MMLVE-Agent consistently achieves high scores across almost all scenes and sub-dimensions, particularly excelling in Identity Preservation (1.1) and Target Execution Accuracy (2.1). 
In contrast, baseline methods like HappyHorse 1.0 and Kling o3 exhibit severe fluctuations and frequent failures (scoring 0 or extremely low marks) in Distractor Isolation (2.2) and Chronological Shot Alignment (3.3), quantitatively reflecting their severe attribute bleeding and temporal scrambling issues. 
Furthermore, Tab.~\ref{tab:addlabel_2} explicitly marks the cases where Seedance 2.0 and Kling o3 rejected processing (denoted by ``/'') due to their internal safety mechanisms when faced with highly complex multi-entity instructions.

\section{Other Results}
\label{sec:other_res}
\begin{figure*}[t]
    \centering
    \includegraphics[width=\linewidth]{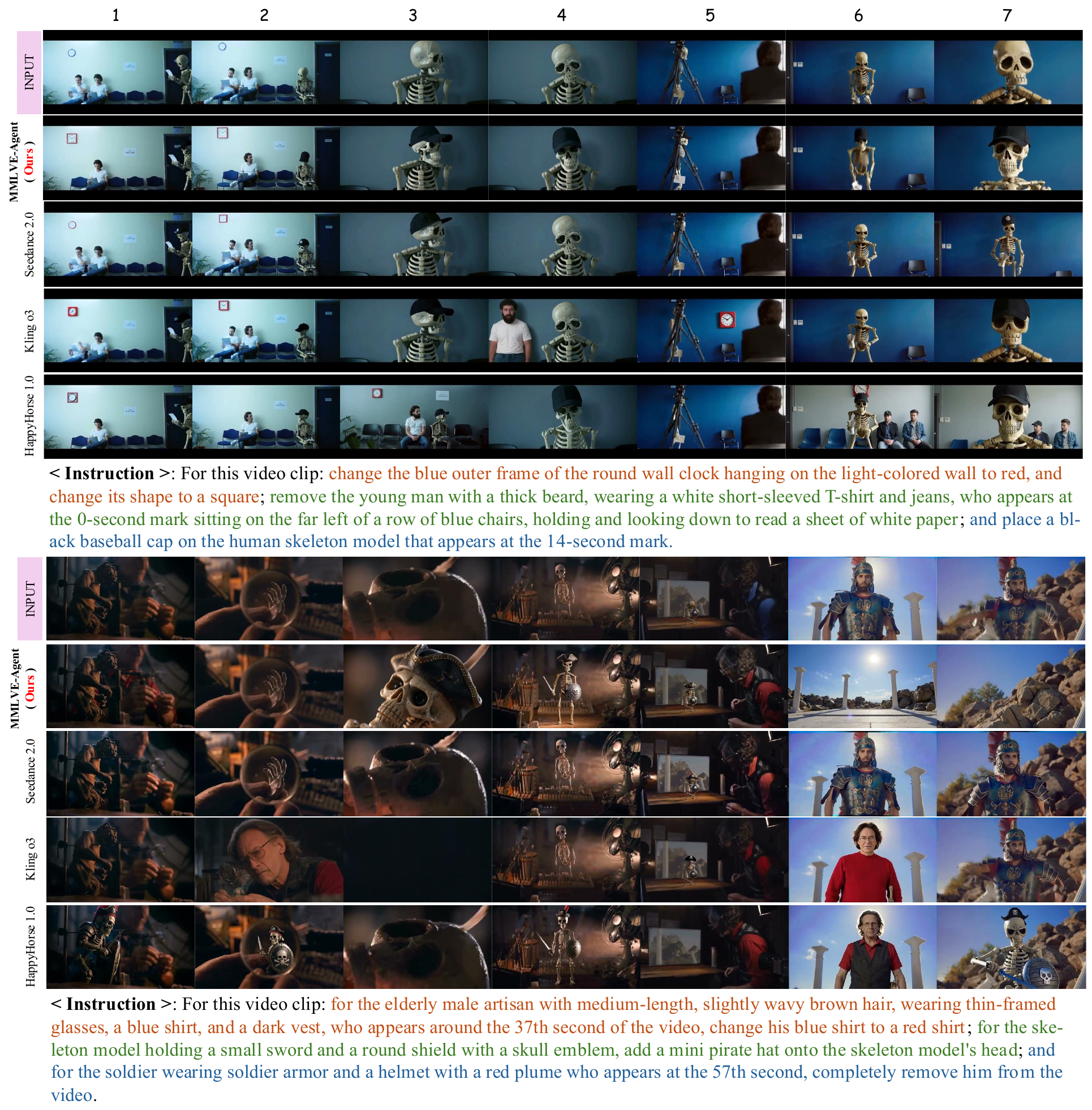}
    \caption{Compared with the baseline methods.}
    \label{fig:supvis4}
\end{figure*}

To further demonstrate the robustness and superiority of our proposed framework, we provide additional qualitative comparisons between MMLVE-Agent and the baseline methods on highly complex multi-shot long videos. 
As shown in Fig.~\ref{fig:supvis4}, the baseline methods consistently struggle with the three core challenges of the MMLVE task, whereas our agentic framework handles them with high precision.

\noindent \textbf{Analysis of the First Case (Top):} 
The first prompt contains three heterogeneous instructions: modifying a clock (shape to square, color to red), deleting a specific man on the left, and adding a cap to a skeleton. 
\textbf{Seedance 2.0} exhibits a highly conservative strategy, leading to severe missed edits (failing MID): it successfully changes the clock's color but fails to alter its shape (Frame 1), completely fails to remove the man (Frames 1-2), and misses the cap addition on the skeleton (Frames 4, 6, 7). 
\textbf{Kling o3} suffers from severe inconsistency and hallucinations: the clock's shape fluctuates (Frames 1-2), and it fails to remove the man. Worse still, it erroneously hallucinates the man into the left side of an unrelated shot (Frame 4) and forcibly inserts the clock into another (Frame 5), while consistently failing to edit the skeleton. 
\textbf{HappyHorse 1.0} demonstrates catastrophic spatiotemporal destruction (failing ZDSS) and instruction interference. While it removes the man, it accidentally deletes the clock as well (Frame 2). It then suffers from severe temporal scrambling (Frame 3) and exhibits extreme hallucinations by forcibly overwriting the original scenes to insert the clock and sitting men into completely unrelated later shots (Frames 6-7). 
In contrast, \textbf{MMLVE-Agent} accurately decouples these instructions, executing all edits flawlessly in their correct temporal windows without any background destruction.

\noindent \textbf{Analysis of the Second Case (Bottom):} 
The second prompt requires changing an artisan's shirt to red, adding a mini pirate hat to a skeleton, and removing a soldier. 
\textbf{Seedance 2.0} again defaults to a conservative failure, completely missing the edits on the skeleton (Frames 3-4) and failing to remove the soldier (Frames 6-7). 
\textbf{Kling o3} suffers from severe chronological misalignment, arbitrarily erasing certain original shots (Frames 2-3, failing ZDSS). Furthermore, it exhibits bizarre attribute bleeding: while it manages to remove the soldier in Frame 6, it forcibly hallucinates the red-shirted artisan into the same scene, and then fails to remove the soldier entirely in Frame 7. 
\textbf{HappyHorse 1.0} displays extreme instruction confusion (failing MID). It erroneously morphs an unrelated object into the pirate-hat skeleton early on (Frames 1-2), fails to edit the actual skeleton (Frame 3), and then bizarrely deletes the skeleton later (Frame 5). In the final shots, although it removes the soldier, it forcibly hallucinates the red-shirted artisan (Frame 6) and the pirate-hat skeleton (Frame 7) into the background. 
Conversely, empowered by the Global Memory Card and the retrieval-based on-demand editing strategy, our \textbf{MMLVE-Agent} strictly isolates the editing operations. It accurately modifies the artisan, edits the skeleton, and removes the soldier exclusively in their respective physical shots, achieving perfect multi-instruction decoupling.

\begin{table*}[t]
  \centering
  \footnotesize
  \resizebox{1.0\textwidth}{!}{
    \begin{tabular}{cccccccccccccccccccc}
    \toprule
    Method & scene & 1.1   & 1.2   & 1.3   & 1.4   & 1.5   & \cellcolor[rgb]{ .949,  .949,  .949}CSEC (1.)  & 2.1   & 2.2   & 2.3   & 2.4   & 2.5   & \cellcolor[rgb]{ .949,  .949,  .949}MID (2.)  & 3.1   & 3.2   & 3.3   & 3.4   & 3.5   & \cellcolor[rgb]{ .949,  .949,  .949}ZDSS (3.) \\
    \midrule
    \multirow{25}[2]{*}{\begin{sideways}MMLVE-Agent\end{sideways}} & sub\_001 & 18    & 19    & 19    & 19    & 19    & \cellcolor[rgb]{ .949,  .949,  .949}94 & 15    & 19    & 20    & 19    & 20    & \cellcolor[rgb]{ .949,  .949,  .949}93 & 18    & 18    & 20    & 19    & 18    & \cellcolor[rgb]{ .949,  .949,  .949}93 \\
          & sub\_002 & 19    & 20    & 19    & 19    & 19    & \cellcolor[rgb]{ .949,  .949,  .949}96 & 20    & 20    & 20    & 20    & 20    & \cellcolor[rgb]{ .949,  .949,  .949}100 & 19    & 20    & 20    & 20    & 19    & \cellcolor[rgb]{ .949,  .949,  .949}98 \\
          & sub\_003 & 20    & 20    & 20    & 20    & 20    & \cellcolor[rgb]{ .949,  .949,  .949}100 & 20    & 20    & 20    & 20    & 20    & \cellcolor[rgb]{ .949,  .949,  .949}100 & 20    & 20    & 20    & 20    & 20    & \cellcolor[rgb]{ .949,  .949,  .949}100 \\
          & sub\_004 & 12    & 14    & 15    & 8     & 10    & \cellcolor[rgb]{ .949,  .949,  .949}59 & 11    & 18    & 6     & 8     & 18    & \cellcolor[rgb]{ .949,  .949,  .949}61 & 12    & 16    & 20    & 10    & 10    & \cellcolor[rgb]{ .949,  .949,  .949}68 \\
          & sub\_005 & 20    & 11    & 19    & 19    & 19    & \cellcolor[rgb]{ .949,  .949,  .949}88 & 14    & 20    & 20    & 19    & 20    & \cellcolor[rgb]{ .949,  .949,  .949}93 & 20    & 20    & 20    & 20    & 19    & \cellcolor[rgb]{ .949,  .949,  .949}99 \\
          & sub\_006 & 20    & 20    & 20    & 19    & 19    & \cellcolor[rgb]{ .949,  .949,  .949}97 & 19    & 20    & 20    & 20    & 20    & \cellcolor[rgb]{ .949,  .949,  .949}99 & 14    & 19    & 20    & 20    & 18    & \cellcolor[rgb]{ .949,  .949,  .949}91 \\
          & sub\_007 & 15    & 18    & 10    & 15    & 15    & \cellcolor[rgb]{ .949,  .949,  .949}73 & 18    & 8     & 15    & 12    & 15    & \cellcolor[rgb]{ .949,  .949,  .949}68 & 5     & 8     & 20    & 15    & 10    & \cellcolor[rgb]{ .949,  .949,  .949}58 \\
          & sub\_008 & 18    & 19    & 19    & 19    & 19    & \cellcolor[rgb]{ .949,  .949,  .949}94 & 19    & 20    & 20    & 19    & 20    & \cellcolor[rgb]{ .949,  .949,  .949}98 & 19    & 19    & 20    & 19    & 19    & \cellcolor[rgb]{ .949,  .949,  .949}96 \\
          & sub\_009 & 20    & 20    & 20    & 20    & 19    & \cellcolor[rgb]{ .949,  .949,  .949}99 & 20    & 19    & 18    & 20    & 20    & \cellcolor[rgb]{ .949,  .949,  .949}97 & 18    & 20    & 20    & 20    & 19    & \cellcolor[rgb]{ .949,  .949,  .949}97 \\
          & sub\_010 & 18    & 20    & 20    & 19    & 19    & \cellcolor[rgb]{ .949,  .949,  .949}96 & 20    & 20    & 20    & 18    & 20    & \cellcolor[rgb]{ .949,  .949,  .949}98 & 20    & 17    & 20    & 20    & 19    & \cellcolor[rgb]{ .949,  .949,  .949}96 \\
          & sub\_011 & 16    & 18    & 18    & 15    & 18    & \cellcolor[rgb]{ .949,  .949,  .949}85 & 20    & 8     & 10    & 10    & 12    & \cellcolor[rgb]{ .949,  .949,  .949}60 & 17    & 16    & 20    & 19    & 18    & \cellcolor[rgb]{ .949,  .949,  .949}90 \\
          & sub\_012 & 19    & 20    & 19    & 19    & 19    & \cellcolor[rgb]{ .949,  .949,  .949}96 & 20    & 20    & 20    & 20    & 20    & \cellcolor[rgb]{ .949,  .949,  .949}100 & 19    & 20    & 20    & 20    & 19    & \cellcolor[rgb]{ .949,  .949,  .949}98 \\
          & sub\_013 & 18    & 18    & 17    & 18    & 17    & \cellcolor[rgb]{ .949,  .949,  .949}88 & 18    & 8     & 15    & 10    & 15    & \cellcolor[rgb]{ .949,  .949,  .949}66 & 12    & 16    & 20    & 19    & 18    & \cellcolor[rgb]{ .949,  .949,  .949}85 \\
          & sub\_014 & 20    & 20    & 20    & 20    & 20    & \cellcolor[rgb]{ .949,  .949,  .949}100 & 20    & 20    & 20    & 20    & 20    & \cellcolor[rgb]{ .949,  .949,  .949}100 & 20    & 20    & 20    & 20    & 20    & \cellcolor[rgb]{ .949,  .949,  .949}100 \\
          & sub\_015 & 18    & 20    & 19    & 18    & 17    & \cellcolor[rgb]{ .949,  .949,  .949}92 & 19    & 19    & 8     & 10    & 18    & \cellcolor[rgb]{ .949,  .949,  .949}74 & 8     & 12    & 20    & 18    & 12    & \cellcolor[rgb]{ .949,  .949,  .949}70 \\
          & sub\_016 & 16    & 14    & 18    & 15    & 17    & \cellcolor[rgb]{ .949,  .949,  .949}80 & 15    & 19    & 17    & 19    & 20    & \cellcolor[rgb]{ .949,  .949,  .949}90 & 15    & 20    & 20    & 19    & 19    & \cellcolor[rgb]{ .949,  .949,  .949}93 \\
          & sub\_017 & 15    & 14    & 16    & 17    & 16    & \cellcolor[rgb]{ .949,  .949,  .949}78 & 15    & 12    & 15    & 13    & 17    & \cellcolor[rgb]{ .949,  .949,  .949}72 & 15    & 17    & 20    & 18    & 16    & \cellcolor[rgb]{ .949,  .949,  .949}86 \\
          & sub\_018 & 16    & 18    & 17    & 16    & 17    & \cellcolor[rgb]{ .949,  .949,  .949}84 & 18    & 20    & 14    & 15    & 20    & \cellcolor[rgb]{ .949,  .949,  .949}87 & 13    & 10    & 20    & 12    & 11    & \cellcolor[rgb]{ .949,  .949,  .949}66 \\
          & sub\_019 & 20    & 20    & 20    & 20    & 18    & \cellcolor[rgb]{ .949,  .949,  .949}98 & 12    & 20    & 12    & 10    & 10    & \cellcolor[rgb]{ .949,  .949,  .949}64 & 12    & 5     & 20    & 6     & 10    & \cellcolor[rgb]{ .949,  .949,  .949}53 \\
          & sub\_020 & 15    & 10    & 16    & 12    & 15    & \cellcolor[rgb]{ .949,  .949,  .949}68 & 8     & 15    & 10    & 9     & 14    & \cellcolor[rgb]{ .949,  .949,  .949}56 & 18    & 18    & 20    & 18    & 14    & \cellcolor[rgb]{ .949,  .949,  .949}88 \\
          & sub\_021 & 16    & 18    & 18    & 17    & 18    & \cellcolor[rgb]{ .949,  .949,  .949}87 & 18    & 5     & 12    & 5     & 8     & \cellcolor[rgb]{ .949,  .949,  .949}48 & 18    & 4     & 20    & 18    & 18    & \cellcolor[rgb]{ .949,  .949,  .949}78 \\
          & sub\_022 & 8     & 12    & 11    & 10    & 12    & \cellcolor[rgb]{ .949,  .949,  .949}53 & 18    & 5     & 8     & 10    & 8     & \cellcolor[rgb]{ .949,  .949,  .949}49 & 10    & 11    & 12    & 10    & 10    & \cellcolor[rgb]{ .949,  .949,  .949}53 \\
          & sub\_023 & 16    & 18    & 15    & 17    & 17    & \cellcolor[rgb]{ .949,  .949,  .949}83 & 13    & 18    & 12    & 10    & 16    & \cellcolor[rgb]{ .949,  .949,  .949}69 & 8     & 10    & 20    & 14    & 10    & \cellcolor[rgb]{ .949,  .949,  .949}62 \\
          & sub\_024 & 18    & 19    & 18    & 18    & 18    & \cellcolor[rgb]{ .949,  .949,  .949}91 & 19    & 20    & 20    & 19    & 20    & \cellcolor[rgb]{ .949,  .949,  .949}98 & 17    & 19    & 20    & 19    & 19    & \cellcolor[rgb]{ .949,  .949,  .949}94 \\
          & sub\_025 & 5     & 8     & 8     & 8     & 12    & \cellcolor[rgb]{ .949,  .949,  .949}41 & 10    & 8     & 4     & 6     & 8     & \cellcolor[rgb]{ .949,  .949,  .949}36 & 8     & 6     & 5     & 5     & 6     & \cellcolor[rgb]{ .949,  .949,  .949}30 \\
    \midrule
    \multirow{25}[2]{*}{\begin{sideways}HappyHorse 1.0\end{sideways}} & sub\_001 & 18    & 17    & 17    & 16    & 17    & \cellcolor[rgb]{ .949,  .949,  .949}85 & 15    & 18    & 12    & 18    & 18    & \cellcolor[rgb]{ .949,  .949,  .949}81 & 18    & 15    & 0     & 10    & 15    & \cellcolor[rgb]{ .949,  .949,  .949}58 \\
          & sub\_002 & 5     & 12    & 13    & 10    & 14    & \cellcolor[rgb]{ .949,  .949,  .949}54 & 14    & 10    & 10    & 8     & 12    & \cellcolor[rgb]{ .949,  .949,  .949}54 & 15    & 11    & 20    & 15    & 14    & \cellcolor[rgb]{ .949,  .949,  .949}75 \\
          & sub\_003 & 18    & 18    & 17    & 18    & 18    & \cellcolor[rgb]{ .949,  .949,  .949}89 & 18    & 16    & 17    & 18    & 19    & \cellcolor[rgb]{ .949,  .949,  .949}88 & 18    & 19    & 20    & 19    & 17    & \cellcolor[rgb]{ .949,  .949,  .949}93 \\
          & sub\_004 & 13    & 14    & 11    & 15    & 12    & \cellcolor[rgb]{ .949,  .949,  .949}65 & 11    & 12    & 8     & 10    & 10    & \cellcolor[rgb]{ .949,  .949,  .949}51 & 12    & 10    & 18    & 15    & 12    & \cellcolor[rgb]{ .949,  .949,  .949}67 \\
          & sub\_005 & 19    & 18    & 19    & 15    & 17    & \cellcolor[rgb]{ .949,  .949,  .949}88 & 19    & 20    & 20    & 16    & 20    & \cellcolor[rgb]{ .949,  .949,  .949}95 & 20    & 20    & 20    & 19    & 16    & \cellcolor[rgb]{ .949,  .949,  .949}95 \\
          & sub\_006 & 14    & 6     & 12    & 15    & 15    & \cellcolor[rgb]{ .949,  .949,  .949}62 & 7     & 20    & 18    & 13    & 20    & \cellcolor[rgb]{ .949,  .949,  .949}78 & 12    & 15    & 20    & 20    & 16    & \cellcolor[rgb]{ .949,  .949,  .949}83 \\
          & sub\_007 & 10    & 4     & 8     & 12    & 10    & \cellcolor[rgb]{ .949,  .949,  .949}44 & 4     & 15    & 12    & 6     & 14    & \cellcolor[rgb]{ .949,  .949,  .949}51 & 4     & 5     & 20    & 10    & 6     & \cellcolor[rgb]{ .949,  .949,  .949}45 \\
          & sub\_008 & 18    & 17    & 18    & 19    & 18    & \cellcolor[rgb]{ .949,  .949,  .949}90 & 12    & 20    & 8     & 8     & 15    & \cellcolor[rgb]{ .949,  .949,  .949}63 & 15    & 4     & 20    & 16    & 15    & \cellcolor[rgb]{ .949,  .949,  .949}70 \\
          & sub\_009 & 20    & 20    & 20    & 19    & 18    & \cellcolor[rgb]{ .949,  .949,  .949}97 & 20    & 20    & 19    & 20    & 20    & \cellcolor[rgb]{ .949,  .949,  .949}99 & 14    & 20    & 20    & 20    & 16    & \cellcolor[rgb]{ .949,  .949,  .949}90 \\
          & sub\_010 & 10    & 10    & 11    & 15    & 14    & \cellcolor[rgb]{ .949,  .949,  .949}60 & 12    & 5     & 6     & 6     & 5     & \cellcolor[rgb]{ .949,  .949,  .949}34 & 4     & 4     & 5     & 8     & 6     & \cellcolor[rgb]{ .949,  .949,  .949}27 \\
          & sub\_011 & 18    & 18    & 19    & 19    & 19    & \cellcolor[rgb]{ .949,  .949,  .949}93 & 20    & 20    & 19    & 20    & 20    & \cellcolor[rgb]{ .949,  .949,  .949}99 & 10    & 8     & 20    & 11    & 12    & \cellcolor[rgb]{ .949,  .949,  .949}61 \\
          & sub\_012 & 10    & 8     & 11    & 12    & 11    & \cellcolor[rgb]{ .949,  .949,  .949}52 & 5     & 4     & 4     & 8     & 6     & \cellcolor[rgb]{ .949,  .949,  .949}27 & 10    & 8     & 14    & 10    & 9     & \cellcolor[rgb]{ .949,  .949,  .949}51 \\
          & sub\_013 & 8     & 10    & 14    & 12    & 15    & \cellcolor[rgb]{ .949,  .949,  .949}59 & 8     & 4     & 3     & 5     & 5     & \cellcolor[rgb]{ .949,  .949,  .949}25 & 18    & 12    & 20    & 14    & 15    & \cellcolor[rgb]{ .949,  .949,  .949}79 \\
          & sub\_014 & 20    & 20    & 19    & 20    & 19    & \cellcolor[rgb]{ .949,  .949,  .949}98 & 20    & 20    & 19    & 18    & 20    & \cellcolor[rgb]{ .949,  .949,  .949}97 & 18    & 15    & 20    & 19    & 18    & \cellcolor[rgb]{ .949,  .949,  .949}90 \\
          & sub\_015 & 14    & 18    & 18    & 15    & 16    & \cellcolor[rgb]{ .949,  .949,  .949}81 & 18    & 18    & 14    & 12    & 18    & \cellcolor[rgb]{ .949,  .949,  .949}80 & 16    & 10    & 5     & 8     & 12    & \cellcolor[rgb]{ .949,  .949,  .949}51 \\
          & sub\_016 & 18    & 20    & 19    & 19    & 20    & \cellcolor[rgb]{ .949,  .949,  .949}96 & 20    & 20    & 18    & 20    & 20    & \cellcolor[rgb]{ .949,  .949,  .949}98 & 19    & 20    & 20    & 20    & 19    & \cellcolor[rgb]{ .949,  .949,  .949}98 \\
          & sub\_017 & 6     & 8     & 10    & 8     & 12    & \cellcolor[rgb]{ .949,  .949,  .949}44 & 8     & 6     & 4     & 5     & 7     & \cellcolor[rgb]{ .949,  .949,  .949}30 & 4     & 3     & 14    & 8     & 6     & \cellcolor[rgb]{ .949,  .949,  .949}35 \\
          & sub\_018 & 12    & 10    & 11    & 12    & 13    & \cellcolor[rgb]{ .949,  .949,  .949}58 & 6     & 14    & 4     & 5     & 10    & \cellcolor[rgb]{ .949,  .949,  .949}39 & 5     & 3     & 10    & 12    & 10    & \cellcolor[rgb]{ .949,  .949,  .949}40 \\
          & sub\_019 & 18    & 19    & 19    & 18    & 18    & \cellcolor[rgb]{ .949,  .949,  .949}92 & 10    & 18    & 15    & 16    & 12    & \cellcolor[rgb]{ .949,  .949,  .949}71 & 16    & 12    & 20    & 8     & 10    & \cellcolor[rgb]{ .949,  .949,  .949}66 \\
          & sub\_020 & 18    & 18    & 18    & 18    & 17    & \cellcolor[rgb]{ .949,  .949,  .949}89 & 8     & 18    & 18    & 18    & 18    & \cellcolor[rgb]{ .949,  .949,  .949}80 & 18    & 18    & 20    & 18    & 18    & \cellcolor[rgb]{ .949,  .949,  .949}92 \\
          & sub\_021 & 19    & 18    & 19    & 20    & 19    & \cellcolor[rgb]{ .949,  .949,  .949}95 & 20    & 20    & 20    & 20    & 20    & \cellcolor[rgb]{ .949,  .949,  .949}100 & 20    & 17    & 20    & 20    & 18    & \cellcolor[rgb]{ .949,  .949,  .949}95 \\
          & sub\_022 & 16    & 18    & 16    & 14    & 15    & \cellcolor[rgb]{ .949,  .949,  .949}79 & 20    & 15    & 13    & 13    & 18    & \cellcolor[rgb]{ .949,  .949,  .949}79 & 9     & 11    & 20    & 11    & 11    & \cellcolor[rgb]{ .949,  .949,  .949}62 \\
          & sub\_023 & 20    & 20    & 20    & 20    & 20    & \cellcolor[rgb]{ .949,  .949,  .949}100 & 10    & 20    & 20    & 15    & 20    & \cellcolor[rgb]{ .949,  .949,  .949}85 & 18    & 12    & 20    & 15    & 16    & \cellcolor[rgb]{ .949,  .949,  .949}81 \\
          & sub\_024 & 10    & 12    & 11    & 10    & 10    & \cellcolor[rgb]{ .949,  .949,  .949}53 & 12    & 6     & 5     & 5     & 8     & \cellcolor[rgb]{ .949,  .949,  .949}36 & 11    & 5     & 12    & 8     & 8     & \cellcolor[rgb]{ .949,  .949,  .949}44 \\
          & sub\_025 & 8     & 8     & 9     & 8     & 11    & \cellcolor[rgb]{ .949,  .949,  .949}44 & 12    & 8     & 6     & 8     & 8     & \cellcolor[rgb]{ .949,  .949,  .949}42 & 5     & 5     & 0     & 6     & 8     & \cellcolor[rgb]{ .949,  .949,  .949}24 \\
    \bottomrule
    \bottomrule
    \end{tabular}%
    }
  \caption{Detail Evaluation Results for MMLVE-Agent and Happyhorse 1.0.}
  \label{tab:addlabel_1}%
\end{table*}%

\begin{table*}[t]
  \centering
  \footnotesize
  \resizebox{1.0\textwidth}{!}{
    \begin{tabular}{cccccccccccccccccccc}
    \toprule
    Method & scene & 1.1   & 1.2   & 1.3   & 1.4   & 1.5   & \cellcolor[rgb]{ .949,  .949,  .949}CSEC (1.)  & 2.1   & 2.2   & 2.3   & 2.4   & 2.5   & \cellcolor[rgb]{ .949,  .949,  .949}MID (2.)  & 3.1   & 3.2   & 3.3   & 3.4   & 3.5   & \cellcolor[rgb]{ .949,  .949,  .949}ZDSS (3.) \\
    \midrule
    \multirow{25}[2]{*}{\begin{sideways}Kling o3\end{sideways}} & sub\_001 & 12    & 18    & 18    & 16    & 18    & \cellcolor[rgb]{ .949,  .949,  .949}82 & 17    & 20    & 20    & 20    & 20    & \cellcolor[rgb]{ .949,  .949,  .949}97 & 5     & 5     & 2     & 8     & 10    & \cellcolor[rgb]{ .949,  .949,  .949}30 \\
          & sub\_002 & /     & /     & /     & /     & /     & \cellcolor[rgb]{ .949,  .949,  .949}/ & /     & /     & /     & /     & /     & \cellcolor[rgb]{ .949,  .949,  .949}/ & /     & /     & /     & /     & /     & \cellcolor[rgb]{ .949,  .949,  .949}/ \\
          & sub\_003 & /     & /     & /     & /     & /     & \cellcolor[rgb]{ .949,  .949,  .949}/ & /     & /     & /     & /     & /     & \cellcolor[rgb]{ .949,  .949,  .949}/ & /     & /     & /     & /     & /     & \cellcolor[rgb]{ .949,  .949,  .949}/ \\
          & sub\_004 & 5     & 5     & 8     & 6     & 8     & \cellcolor[rgb]{ .949,  .949,  .949}32 & 5     & 6     & 5     & 4     & 10    & \cellcolor[rgb]{ .949,  .949,  .949}30 & 10    & 12    & 20    & 8     & 10    & \cellcolor[rgb]{ .949,  .949,  .949}60 \\
          & sub\_005 & 14    & 8     & 12    & 14    & 12    & \cellcolor[rgb]{ .949,  .949,  .949}60 & 10    & 15    & 15    & 15    & 15    & \cellcolor[rgb]{ .949,  .949,  .949}70 & 10    & 10    & 5     & 12    & 12    & \cellcolor[rgb]{ .949,  .949,  .949}49 \\
          & sub\_006 & 11    & 9     & 12    & 10    & 8     & \cellcolor[rgb]{ .949,  .949,  .949}50 & 8     & 6     & 5     & 10    & 7     & \cellcolor[rgb]{ .949,  .949,  .949}36 & 5     & 4     & 2     & 8     & 8     & \cellcolor[rgb]{ .949,  .949,  .949}27 \\
          & sub\_007 & 10    & 8     & 12    & 12    & 11    & \cellcolor[rgb]{ .949,  .949,  .949}53 & 10    & 8     & 2     & 4     & 8     & \cellcolor[rgb]{ .949,  .949,  .949}32 & 0     & 0     & 0     & 4     & 4     & \cellcolor[rgb]{ .949,  .949,  .949}8 \\
          & sub\_008 & 13    & 14    & 8     & 11    & 7     & \cellcolor[rgb]{ .949,  .949,  .949}53 & 13    & 12    & 6     & 8     & 12    & \cellcolor[rgb]{ .949,  .949,  .949}51 & 7     & 8     & 15    & 10    & 6     & \cellcolor[rgb]{ .949,  .949,  .949}46 \\
          & sub\_009 & 15    & 16    & 16    & 15    & 12    & \cellcolor[rgb]{ .949,  .949,  .949}74 & 10    & 18    & 18    & 10    & 18    & \cellcolor[rgb]{ .949,  .949,  .949}74 & 14    & 13    & 20    & 15    & 15    & \cellcolor[rgb]{ .949,  .949,  .949}77 \\
          & sub\_010 & 8     & 6     & 8     & 10    & 10    & \cellcolor[rgb]{ .949,  .949,  .949}42 & 10    & 18    & 14    & 12    & 14    & \cellcolor[rgb]{ .949,  .949,  .949}68 & 4     & 4     & 0     & 10    & 8     & \cellcolor[rgb]{ .949,  .949,  .949}26 \\
          & sub\_011 & 20    & 10    & 18    & 20    & 19    & \cellcolor[rgb]{ .949,  .949,  .949}87 & 10    & 20    & 20    & 20    & 20    & \cellcolor[rgb]{ .949,  .949,  .949}90 & 18    & 19    & 20    & 20    & 18    & \cellcolor[rgb]{ .949,  .949,  .949}95 \\
          & sub\_012 & 10    & 10    & 12    & 12    & 14    & \cellcolor[rgb]{ .949,  .949,  .949}58 & 8     & 6     & 10    & 7     & 10    & \cellcolor[rgb]{ .949,  .949,  .949}41 & 18    & 10    & 20    & 16    & 15    & \cellcolor[rgb]{ .949,  .949,  .949}79 \\
          & sub\_013 & 15    & 18    & 16    & 10    & 12    & \cellcolor[rgb]{ .949,  .949,  .949}71 & 14    & 6     & 10    & 8     & 15    & \cellcolor[rgb]{ .949,  .949,  .949}53 & 18    & 16    & 20    & 18    & 11    & \cellcolor[rgb]{ .949,  .949,  .949}83 \\
          & sub\_014 & 20    & 20    & 20    & 20    & 19    & \cellcolor[rgb]{ .949,  .949,  .949}98 & 20    & 20    & 20    & 20    & 20    & \cellcolor[rgb]{ .949,  .949,  .949}100 & 20    & 18    & 20    & 20    & 20    & \cellcolor[rgb]{ .949,  .949,  .949}98 \\
          & sub\_015 & 20    & 19    & 19    & 20    & 20    & \cellcolor[rgb]{ .949,  .949,  .949}98 & 18    & 20    & 20    & 20    & 20    & \cellcolor[rgb]{ .949,  .949,  .949}98 & 16    & 20    & 20    & 20    & 20    & \cellcolor[rgb]{ .949,  .949,  .949}96 \\
          & sub\_016 & 18    & 10    & 18    & 19    & 18    & \cellcolor[rgb]{ .949,  .949,  .949}83 & 12    & 20    & 20    & 20    & 20    & \cellcolor[rgb]{ .949,  .949,  .949}92 & 19    & 20    & 20    & 20    & 19    & \cellcolor[rgb]{ .949,  .949,  .949}98 \\
          & sub\_017 & 6     & 8     & 10    & 10    & 12    & \cellcolor[rgb]{ .949,  .949,  .949}46 & 10    & 6     & 8     & 10    & 12    & \cellcolor[rgb]{ .949,  .949,  .949}46 & 4     & 5     & 8     & 8     & 6     & \cellcolor[rgb]{ .949,  .949,  .949}31 \\
          & sub\_018 & 19    & 19    & 18    & 19    & 19    & \cellcolor[rgb]{ .949,  .949,  .949}94 & 19    & 20    & 20    & 20    & 20    & \cellcolor[rgb]{ .949,  .949,  .949}99 & 20    & 20    & 20    & 20    & 19    & \cellcolor[rgb]{ .949,  .949,  .949}99 \\
          & sub\_019 & 0     & 0     & 0     & 0     & 0     & \cellcolor[rgb]{ .949,  .949,  .949}0 & 2     & 20    & 15    & 5     & 10    & \cellcolor[rgb]{ .949,  .949,  .949}52 & 15    & 8     & 20    & 4     & 8     & \cellcolor[rgb]{ .949,  .949,  .949}55 \\
          & sub\_020 & 18    & 13    & 13    & 18    & 18    & \cellcolor[rgb]{ .949,  .949,  .949}80 & 15    & 19    & 19    & 19    & 19    & \cellcolor[rgb]{ .949,  .949,  .949}91 & 20    & 20    & 20    & 19    & 19    & \cellcolor[rgb]{ .949,  .949,  .949}98 \\
          & sub\_021 & 19    & 19    & 19    & 19    & 19    & \cellcolor[rgb]{ .949,  .949,  .949}95 & 19    & 20    & 14    & 19    & 20    & \cellcolor[rgb]{ .949,  .949,  .949}92 & 19    & 19    & 20    & 19    & 18    & \cellcolor[rgb]{ .949,  .949,  .949}95 \\
          & sub\_022 & 19    & 14    & 18    & 18    & 18    & \cellcolor[rgb]{ .949,  .949,  .949}87 & 12    & 20    & 19    & 19    & 20    & \cellcolor[rgb]{ .949,  .949,  .949}90 & 19    & 20    & 20    & 19    & 19    & \cellcolor[rgb]{ .949,  .949,  .949}97 \\
          & sub\_023 & 20    & 20    & 19    & 19    & 18    & \cellcolor[rgb]{ .949,  .949,  .949}96 & 14    & 20    & 20    & 20    & 20    & \cellcolor[rgb]{ .949,  .949,  .949}94 & 20    & 20    & 20    & 20    & 20    & \cellcolor[rgb]{ .949,  .949,  .949}100 \\
          & sub\_024 & 20    & 20    & 20    & 20    & 20    & \cellcolor[rgb]{ .949,  .949,  .949}100 & 20    & 20    & 20    & 20    & 20    & \cellcolor[rgb]{ .949,  .949,  .949}100 & 20    & 20    & 20    & 20    & 20    & \cellcolor[rgb]{ .949,  .949,  .949}100 \\
          & sub\_025 & 10    & 12    & 15    & 16    & 18    & \cellcolor[rgb]{ .949,  .949,  .949}71 & 14    & 18    & 5     & 18    & 18    & \cellcolor[rgb]{ .949,  .949,  .949}73 & 10    & 8     & 5     & 10    & 10    & \cellcolor[rgb]{ .949,  .949,  .949}43 \\
    \midrule
    \multirow{25}[2]{*}{\begin{sideways}Seedance 2.0\end{sideways}} & sub\_001 & 20    & 20    & 19    & 20    & 19    & \cellcolor[rgb]{ .949,  .949,  .949}98 & 20    & 20    & 20    & 20    & 20    & \cellcolor[rgb]{ .949,  .949,  .949}100 & 20    & 20    & 20    & 20    & 20    & \cellcolor[rgb]{ .949,  .949,  .949}100 \\
          & sub\_002 & 11    & 10    & 14    & 12    & 13    & \cellcolor[rgb]{ .949,  .949,  .949}60 & 13    & 16    & 15    & 15    & 17    & \cellcolor[rgb]{ .949,  .949,  .949}76 & 5     & 4     & 5     & 9     & 7     & \cellcolor[rgb]{ .949,  .949,  .949}30 \\
          & sub\_003 & 18    & 20    & 18    & 18    & 18    & \cellcolor[rgb]{ .949,  .949,  .949}92 & 20    & 20    & 10    & 19    & 20    & \cellcolor[rgb]{ .949,  .949,  .949}89 & 18    & 12    & 20    & 18    & 17    & \cellcolor[rgb]{ .949,  .949,  .949}85 \\
          & sub\_004 & 12    & 10    & 11    & 8     & 9     & \cellcolor[rgb]{ .949,  .949,  .949}50 & 10    & 15    & 12    & 10    & 12    & \cellcolor[rgb]{ .949,  .949,  .949}59 & 12    & 10    & 5     & 12    & 10    & \cellcolor[rgb]{ .949,  .949,  .949}49 \\
          & sub\_005 & 20    & 20    & 20    & 20    & 20    & \cellcolor[rgb]{ .949,  .949,  .949}100 & 20    & 20    & 20    & 20    & 20    & \cellcolor[rgb]{ .949,  .949,  .949}100 & 20    & 20    & 20    & 20    & 20    & \cellcolor[rgb]{ .949,  .949,  .949}100 \\
          & sub\_006 & 12    & 5     & 8     & 10    & 8     & \cellcolor[rgb]{ .949,  .949,  .949}43 & 5     & 12    & 10    & 5     & 8     & \cellcolor[rgb]{ .949,  .949,  .949}40 & 8     & 4     & 0     & 10    & 12    & \cellcolor[rgb]{ .949,  .949,  .949}34 \\
          & sub\_007 & /     & /     & /     & /     & /     & \cellcolor[rgb]{ .949,  .949,  .949}/ & /     & /     & /     & /     & /     & \cellcolor[rgb]{ .949,  .949,  .949}/ & /     & /     & /     & /     & /     & \cellcolor[rgb]{ .949,  .949,  .949}/ \\
          & sub\_008 & 18    & 18    & 18    & 18    & 18    & \cellcolor[rgb]{ .949,  .949,  .949}90 & 18    & 19    & 19    & 18    & 19    & \cellcolor[rgb]{ .949,  .949,  .949}93 & 18    & 18    & 20    & 19    & 18    & \cellcolor[rgb]{ .949,  .949,  .949}93 \\
          & sub\_009 & 20    & 20    & 19    & 18    & 17    & \cellcolor[rgb]{ .949,  .949,  .949}94 & 10    & 20    & 20    & 20    & 20    & \cellcolor[rgb]{ .949,  .949,  .949}90 & 20    & 20    & 20    & 20    & 20    & \cellcolor[rgb]{ .949,  .949,  .949}100 \\
          & sub\_010 & 8     & 6     & 8     & 10    & 12    & \cellcolor[rgb]{ .949,  .949,  .949}44 & 8     & 2     & 2     & 4     & 5     & \cellcolor[rgb]{ .949,  .949,  .949}21 & 14    & 2     & 20    & 12    & 12    & \cellcolor[rgb]{ .949,  .949,  .949}60 \\
          & sub\_011 & 20    & 20    & 20    & 20    & 19    & \cellcolor[rgb]{ .949,  .949,  .949}99 & 20    & 20    & 20    & 20    & 20    & \cellcolor[rgb]{ .949,  .949,  .949}100 & 20    & 19    & 20    & 20    & 20    & \cellcolor[rgb]{ .949,  .949,  .949}99 \\
          & sub\_012 & 0     & 0     & 0     & 0     & 0     & \cellcolor[rgb]{ .949,  .949,  .949}0 & 0     & 5     & 5     & 0     & 0     & \cellcolor[rgb]{ .949,  .949,  .949}10 & 5     & 2     & 15    & 5     & 5     & \cellcolor[rgb]{ .949,  .949,  .949}32 \\
          & sub\_013 & 13    & 12    & 12    & 16    & 15    & \cellcolor[rgb]{ .949,  .949,  .949}68 & 18    & 9     & 14    & 10    & 15    & \cellcolor[rgb]{ .949,  .949,  .949}66 & 18    & 11    & 20    & 16    & 11    & \cellcolor[rgb]{ .949,  .949,  .949}76 \\
          & sub\_014 & 20    & 20    & 20    & 20    & 19    & \cellcolor[rgb]{ .949,  .949,  .949}99 & 20    & 20    & 20    & 20    & 20    & \cellcolor[rgb]{ .949,  .949,  .949}100 & 19    & 20    & 20    & 20    & 20    & \cellcolor[rgb]{ .949,  .949,  .949}99 \\
          & sub\_015 & 18    & 18    & 17    & 18    & 17    & \cellcolor[rgb]{ .949,  .949,  .949}88 & 12    & 18    & 18    & 17    & 18    & \cellcolor[rgb]{ .949,  .949,  .949}83 & 17    & 15    & 20    & 18    & 15    & \cellcolor[rgb]{ .949,  .949,  .949}85 \\
          & sub\_016 & 14    & 8     & 14    & 15    & 14    & \cellcolor[rgb]{ .949,  .949,  .949}65 & 7     & 18    & 19    & 18    & 18    & \cellcolor[rgb]{ .949,  .949,  .949}80 & 19    & 19    & 20    & 18    & 18    & \cellcolor[rgb]{ .949,  .949,  .949}94 \\
          & sub\_017 & 10    & 8     & 10    & 12    & 14    & \cellcolor[rgb]{ .949,  .949,  .949}54 & 6     & 15    & 16    & 12    & 12    & \cellcolor[rgb]{ .949,  .949,  .949}61 & 18    & 16    & 20    & 18    & 15    & \cellcolor[rgb]{ .949,  .949,  .949}87 \\
          & sub\_018 & 18    & 19    & 18    & 18    & 17    & \cellcolor[rgb]{ .949,  .949,  .949}90 & 19    & 20    & 20    & 18    & 20    & \cellcolor[rgb]{ .949,  .949,  .949}97 & 14    & 17    & 20    & 18    & 14    & \cellcolor[rgb]{ .949,  .949,  .949}83 \\
          & sub\_019 & 19    & 18    & 18    & 18    & 18    & \cellcolor[rgb]{ .949,  .949,  .949}91 & 10    & 20    & 20    & 20    & 18    & \cellcolor[rgb]{ .949,  .949,  .949}88 & 19    & 19    & 20    & 19    & 19    & \cellcolor[rgb]{ .949,  .949,  .949}96 \\
          & sub\_020 & 18    & 18    & 18    & 18    & 18    & \cellcolor[rgb]{ .949,  .949,  .949}90 & 8     & 18    & 18    & 18    & 18    & \cellcolor[rgb]{ .949,  .949,  .949}80 & 19    & 19    & 20    & 19    & 19    & \cellcolor[rgb]{ .949,  .949,  .949}96 \\
          & sub\_021 & 20    & 17    & 17    & 19    & 18    & \cellcolor[rgb]{ .949,  .949,  .949}91 & 19    & 20    & 20    & 19    & 20    & \cellcolor[rgb]{ .949,  .949,  .949}98 & 19    & 20    & 20    & 20    & 19    & \cellcolor[rgb]{ .949,  .949,  .949}98 \\
          & sub\_022 & 12    & 10    & 12    & 14    & 15    & \cellcolor[rgb]{ .949,  .949,  .949}63 & 10    & 12    & 16    & 10    & 15    & \cellcolor[rgb]{ .949,  .949,  .949}63 & 18    & 12    & 20    & 14    & 15    & \cellcolor[rgb]{ .949,  .949,  .949}79 \\
          & sub\_023 & 20    & 20    & 20    & 20    & 19    & \cellcolor[rgb]{ .949,  .949,  .949}99 & 12    & 20    & 20    & 20    & 20    & \cellcolor[rgb]{ .949,  .949,  .949}92 & 20    & 20    & 20    & 20    & 20    & \cellcolor[rgb]{ .949,  .949,  .949}100 \\
          & sub\_024 & 20    & 20    & 20    & 20    & 20    & \cellcolor[rgb]{ .949,  .949,  .949}100 & 20    & 20    & 20    & 20    & 20    & \cellcolor[rgb]{ .949,  .949,  .949}100 & 20    & 20    & 20    & 20    & 20    & \cellcolor[rgb]{ .949,  .949,  .949}100 \\
          & sub\_025 & 19    & 19    & 19    & 19    & 18    & \cellcolor[rgb]{ .949,  .949,  .949}94 & 20    & 20    & 20    & 20    & 20    & \cellcolor[rgb]{ .949,  .949,  .949}100 & 20    & 20    & 20    & 20    & 19    & \cellcolor[rgb]{ .949,  .949,  .949}99 \\
    \bottomrule
    \bottomrule
    \end{tabular}%
    }
  \caption{Detail Evaluation Results for Kling o3 and Seedance 2.0. ``$/$'' means the method rejects processing due to its safety mechanism for the scene.}
  \label{tab:addlabel_2}%
\end{table*}%
\end{document}